\documentclass[10pt,twocolumn,letterpaper]{article}

\usepackage{cvpr}              
\definecolor{cvprblue}{rgb}{0.21,0.49,0.74}
\usepackage[pagebackref,breaklinks,colorlinks,allcolors=cvprblue]{hyperref}

\usepackage{amsmath}
\usepackage{graphicx}
\usepackage{multirow}
\usepackage{wrapfig}
\usepackage{booktabs}
\usepackage[toc,page,header]{appendix}
\usepackage{minitoc}
\usepackage[accsupp]{axessibility}  

\def\dataset/{Sim2Struct-1000}
\def\method/{CryoHype}

\def\paperID{} 
\def\confName{CVPR}
\def\confYear{2026}

\title{CryoHype: Reconstructing a thousand cryo-EM structures with transformer-based hypernetworks}

\author{\textbf{Jeffrey Gu}$^{1}$\thanks{Equal contribution. JG completed the initial part of this work while at Stanford University.}~, \textbf{Minkyu Jeon}$^{1}$\footnotemark[\value{footnote}]~, \textbf{Ambri Ma}$^1$, \textbf{Serena Yeung-Levy}$^2$, \textbf{Ellen D. Zhong}$^1$ \\
$^1$Princeton University \quad $^2$Stanford University
}

\begin{document}
\maketitle

\doparttoc 
\faketableofcontents 
\part{} 

\begin{abstract}
Cryo-electron microscopy (cryo-EM) is an indispensable technique for determining the 3D structures of dynamic biomolecular complexes. While typically applied to image a single molecular species, cryo-EM has the potential for structure determination of many targets simultaneously in a high-throughput fashion. However, existing methods typically focus on modeling conformational heterogeneity within a single or a few structures and are not designed to resolve compositional heterogeneity arising from mixtures of many distinct molecular species. To address this challenge, we propose \method/, a transformer-based hypernetwork for cryo-EM reconstruction that dynamically adjusts the weights of an implicit neural representation. Using \method/, we achieve state-of-the-art results on a challenging benchmark dataset containing 100 structures. We further demonstrate that \method/ scales to the reconstruction of 1,000 distinct structures from unlabeled cryo-EM images in the fixed-pose setting. Project page: \url{https://cryohype.cs.princeton.edu/}.

\end{abstract}
\section{Introduction}
\label{sec:intro}

Single particle cryo-electron microscopy (cryo-EM) has emerged as an essential tool to resolve the 3D structures of macromolecular complexes at atomic resolution~\citep{cheng2018single,nakane2018characterisation, yip2020atomic}. Unlike static structure prediction algorithms or other structure determination methods, cryo-EM can experimentally probe the dynamic conformational states of large macromolecular complexes. However, 3D reconstruction of the resulting images poses a challenging inverse problem. 


Although cryo-EM is typically used to resolve \textit{conformational heterogeneity} within a single or a few structures from a purified sample, the technique is increasingly being used to capture more complex scenarios, including heterogeneous mixtures, unpurified samples, or cellular lysates~\citep{ho2020bottom,rabuck2022quantitative,jeon2024cryobench,nogales2024bridging}. 
The ability to simultaneously resolve multiple distinct structures (\emph{compositional heterogeneity}) presents a major opportunity for high-throughput structural discovery. However, the complexity of these mixtures poses new modeling and inference challenges for cryo-EM reconstruction. In this work, we develop new architectures for modeling \textit{extreme} compositional heterogeneity that can scale to datasets containing 10s-100s of distinct structures.

The standard approach for resolving discrete heterogeneity is 3D classification~\citep{scheres2007disentangling, scheres2012relion, scheres2016processing, punjani2017cryosparc, grant2018cis}, a formulation that is naturally suited for modeling mixtures of distinct molecular species. 
While 3D classification models the dataset as originating from a discrete mixture of $K$ independent voxel arrays, its reliance on Expectation-Maximization (EM) for inference limits its scalability to a small number of classes (typically $K < 10$). Because computational cost and memory requirements scale linearly with $K$, marginalization over class assignments (and optionally poses) becomes intractable for large values of $K$. Furthermore, in this regime, EM is highly sensitive to initialization and prone to optimization instability, often resulting in mode collapse and empty clusters.

In contrast, recent deep learning methods based on implicit or explicit neural representations~\citep{zhongreconstructing, zhong2021cryodrgn, kimanius2022sparse, levy2024mixture, levy2025cryodrgn, herreros2025real, qu2025cryonerf} bypass the need for discrete classes by learning a continuous mapping from a continuous latent space to the volume density. While effective for modeling conformational heterogeneity, these methods are ill-suited for extreme compositional heterogeneity. Most architectures generate volumes using a fixed decoder conditioned on a latent vector. This design forces morphologically distinct structures to be represented by a single set of shared network weights. When applied to hundreds of distinct species, this excessive parameter sharing limits the model's capacity to resolve high-frequency details for any individual structure.


\begin{figure*}[htb!]
  \centering
  \includegraphics[width=0.9\linewidth]{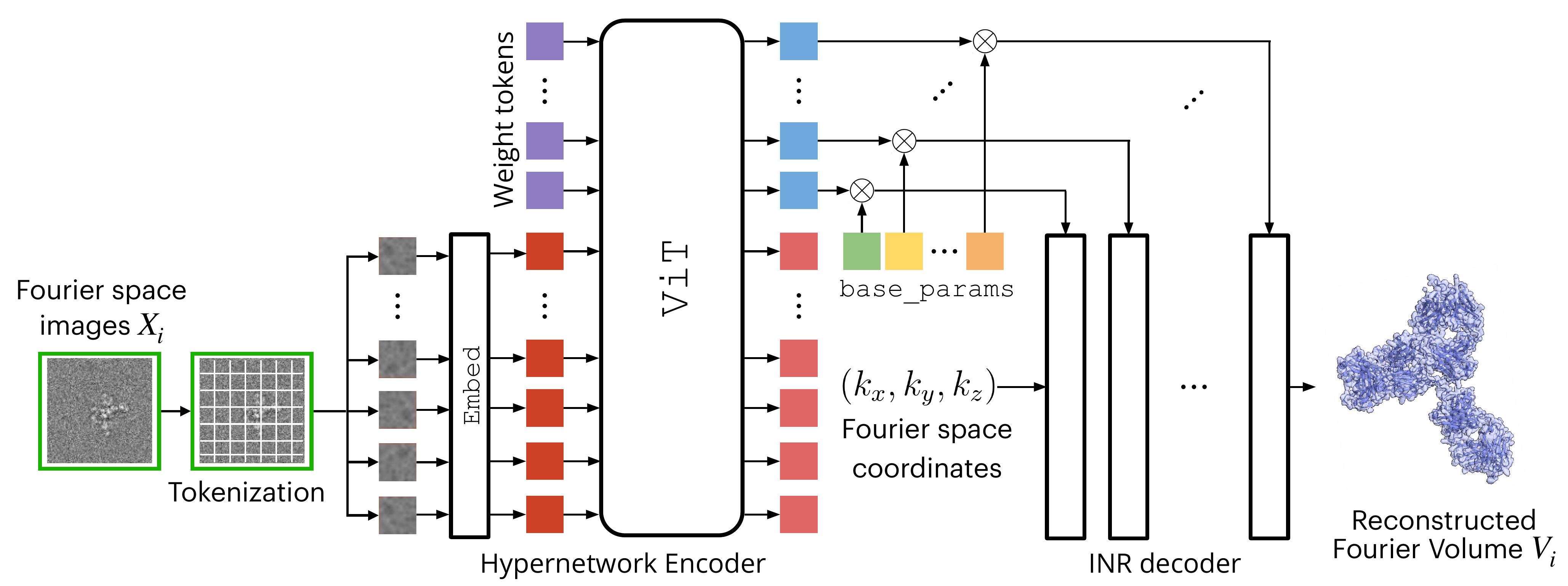}
  \caption{\textbf{CryoHype architecture}. An input image $X_i$ of an unknown structure is first tokenized and concatenated with learnable weight tokens. All tokens are then processed with a transformer encoder, and the output weight tokens are used to modify the weights of an implicit neural representation (INR) that reconstructs the structure $V_i$.}
  \label{fig:framework}
\end{figure*}

Here, we introduce \method/, a transformer-based hypernetwork architecture designed to resolve extreme compositional heterogeneity. Using a hypernetwork~\citep{ha2016hypernetworks} encoder allows the model to dynamically adapt the weights of the neural representation to each distinct structure, reducing parameter sharing and increasing expressivity compared to conditioning by concatenation~\citep{sitzmann2020metasdf, dumoulin2018feature, mehta2021modulated}. The vision transformer~\citep{dosovitskiy2020image} (ViT) architecture for the hypernetwork provides scalable and parameter-efficient weight generation. In order to evaluate \method/, we introduce \texttt{\dataset/}, a dataset for extreme compositional heterogeneity containing a mixture of 1,000 distinct structures, 10 times more structures than previous datasets for compositional heterogeneity~\citep{jeon2024cryobench}. Since traditional resolution-based metrics such as Fourier Shell Correlation (FSC) can be insensitive to variations in shape~\citep{gilles2025cryo}, we additionally evaluate our method using real-space 3D shape metrics. We demonstrate state-of-the-art performance with a 67\% increase in $\mathrm{FSC}_{\mathrm{AUC}}$ on the \texttt{Tomotwin-100} dataset over the best baseline~\citep{jeon2024cryobench}. We further show that the hypernetwork architecture can scale to extreme compositional heterogeneity by reconstructing 1,000 unknown structures in the fixed-pose setting. 

\section{Methods}
\label{sec:methods}

In this section, we introduce the cryo-EM image formation model (Section~\ref{sec:image-formation}), motivation (Section~\ref{sec:methods-motivation}), and our transformer-based hypernetwork method, \method/ (Section~\ref{sec:vit-hypernet}). 

\subsection{Cryo-EM Image Formation Model}
\label{sec:image-formation}

The cryo-EM reconstruction task is to recover density volumes $V_i: \mathbb{R}^3 \to \mathbb{R}, 1 \le i \le N$ representing the electron scattering potential of a molecule of interest from a set of noisy 2D projection images $X_1, \ldots, X_N$ of $V_i$. In each projection $X_i$, the molecule is captured in an unknown pose $\phi_i=(R_i,t_i)$ relative to the electron beam, consisting of a rotation $R_i \in SO(3)$ and in-plane translation $t_i \in \mathbb{R}^2$. In Fourier space, the image formation model can be written as:
\begin{align}
\label{eqn:real-img-form}
    X_i = C_i\mathcal{P}_{\phi_i} V_i + \epsilon_i
\end{align}
where $C_i$ is the Contrast Transfer Function (CTF), $\mathcal{P}_{\phi_i}$ is a slicing operator corresponding to projecting $V_i$ at pose $\phi_i$, and $\epsilon_i \sim \mathcal{N}(0, \sigma^2)$ models additive isotropic Gaussian noise. Homogeneous reconstruction algorithms assume a single density volume $V$ in the image formation model, whereas heterogeneous reconstruction algorithms assume that the volumes $\{V_i\}$ are independent samples from a distribution $\mathbb{P}_V$ over the space of all possible 3D potentials. Additional details are provided in Appendix~\ref{sec:si-image-formation}.

\subsection{Motivation}
\label{sec:methods-motivation}

Existing neural methods for heterogeneous cryo-EM reconstruction condition a shared neural volume representation on a latent code, either as an additional input to an implicit neural representation (INR)~\citep{zhongreconstructing} or as the coefficients of a linear combination of a basis of voxel arrays~\citep{kimanius2022sparse} (see Section~\ref{sec:related-work}). In these approaches, almost all parameters of the neural volume representations are shared among all the different structures, limiting the diversity of the structures that can be captured and the ability of the model to generate structure-specific high-resolution details. Hypernetworks overcome this problem by increasing the expressiveness of conditioning~\cite{galanti2020modularity} and reducing parameter sharing between decoders, since it can be proven that conditioning a network $\Psi$ by concatenation is equivalent to having a linear hypernetwork produce the biases of the first layer of $\Psi$~\citep{sitzmann2020metasdf, dumoulin2018feature, mehta2021modulated}.
Previous work has shown that the first layer is not the optimal layer for weight modulation~\cite{kim2023generalizable, vyas2024learning} and modulating all layers outperforms adjusting a single layer~\cite{gu2025foundation}. Therefore, a full hypernetwork generalizes conditioning by concatenation and can be much more expressive, especially if the hypernetwork is more expressive than a linear layer and significantly increases the proportion of non-shared parameters than previous approaches.  
\begin{figure}
  \centering
  \includegraphics[width=0.9\linewidth]{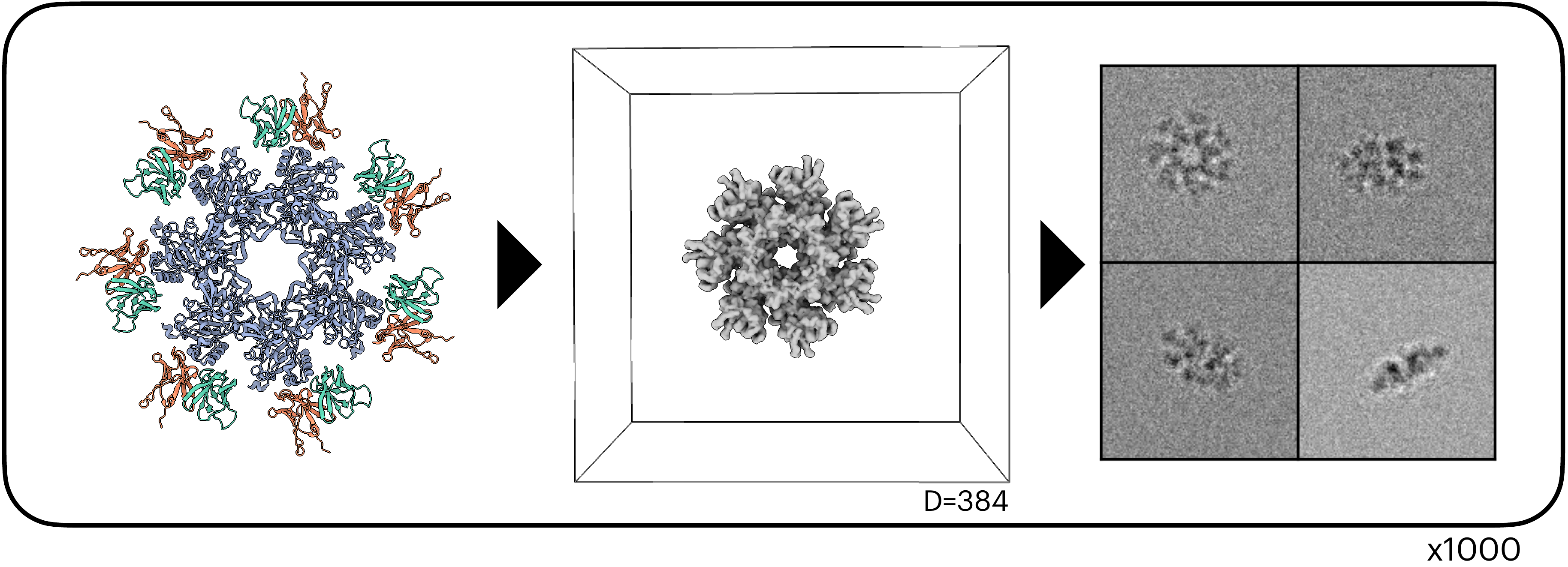}
  \caption{\textbf{\dataset/.} Example atomic model, density map, and projected images from \texttt{\dataset/}, containing 1000 distinct structures. }
  \label{fig:ambris-dataset}
  \vspace{-10pt}
\end{figure}
\begin{figure}
  \centering
  \includegraphics[width=0.9\linewidth]{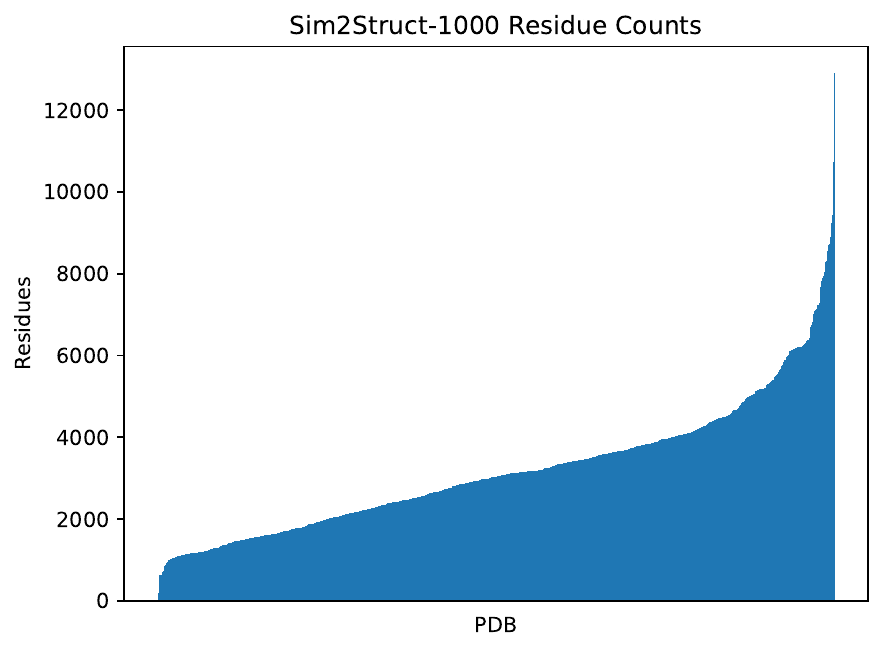}
  \caption{\textbf{\dataset/.} Residue counts of all 1000 structures in \texttt{Sim2Struct-1000}.}
  \label{fig:s2s-residue-counts}
  \vspace{-15pt}
\end{figure}

\subsection{\method/ architecture}
\label{sec:vit-hypernet}

The \method/ architecture consists of five main components: (1) a ViT encoder $g$, consisting of a tokenizer \texttt{Embed} and Transformer encoder \texttt{Enc}, (2) extra learnable weight tokens $\{w_i\}_{i = 1}^q$ (3) an INR $f$, a ReLU MLP with residual connections, with a shared set of base parameters $\{\theta^j\}_{j=1}^{L}$ where $L$ is the number of layers, and (4) learnable linear heads $\{\texttt{Head}_j\}_{j = 1}^{L}$ for each layer $L_j$ in $f$ (see Figure \ref{fig:framework}). Reconstruction is done completely in the Fourier domain. A forward pass of our model works as follows: first, an input projection $\hat{X}$ tokenized into $T$ tokens $\{t_k\}_{k = 1}^T$ by $\texttt{Embed}$. These $T$ tokens are then concatenated along with the learnable weight tokens $w_i$ and processed by \texttt{Enc}, 
the Transformer part of the ViT encoder, to produce the final tokens $[t_1^F, \ldots, t_{T}^F, w_1^F, \ldots, w_q^F]$. The final output tokens $\{w_i^F\}_{i = 1}^q$ are then divided into $L$ groups consisting of $a_j$ tokens $w_1^{F, j}, \ldots, w_{a_j}^{F, j}, 1 \le j \le L$, with $\sum_j a_j = q$. The $j$th group $[w_{1}^{F, j}, \ldots, w_{a_j}^{F, j}]$ is transformed by the linear head $\texttt{Head}_j$ and normalized. The output of the previous step is multiplied elementwise by the $j$th layer's base parameter $\theta_j$ to produce the final parameters $\theta_j^F$ of the $j$th layer:
\begin{align}
    \theta_j^F = \texttt{Norm}(\texttt{Head}_j([w_{1}^{F, j}, \ldots, w_{a_j}^{F, j}])) \otimes \theta_j
\end{align} 

The final INR parameters $\theta_j^F$ are used to instantiate the INR $f$, which parametrizes the structure $\hat{V}$. The INR $f$ maps Fourier space coordinates $(k_x, k_y, k_z)$ to the Fourier-transformed electron scattering potential at that coordinate, producing a clean (i.e., not noisy and CTF-free) prediction $\Tilde{X}$. $\Tilde{X}$ is then multiplied by the CTF (see Section~\ref{sec:image-formation}), and a reconstruction loss (mean-square error, MSE) is computed between the ground truth views and predicted views, and gradients are backpropagated to the hypernetwork. Note that \method/ is trained end-to-end, with the learnable parameters being (1) the ViT encoder $g$, (2) the extra learnable weight tokens $w_i, 1 \le i \le q$, (3) the decoder's base parameters $\theta_j, 1 \le j \le L$, where $L$ is the number of layers in the decoder, and (4) the learnable linear heads $\texttt{Head}_j, 1 \le j \le L$.

\textbf{Latent space embeddings.} Unlike autoencoder and autodecoder-based reconstruction methods, \method/ does not have a canonical low-dimensional latent space. For our latent space analysis, we consider the set of tokens $\{w_i^F\}_{i = 1}^F$ (blue square in Fig.~\ref{fig:framework}) to comprise the model's latent space. These tokens have total dimension $qd$ where $q$ is the number of weight tokens and $d$ is the dimension of the ViT, with the total latent dimension being extremely high-dimensional. To get an interpretable latent space, we perform dimensionality reduction in two stages: first, we use principal component analysis (PCA) to reduce to a smaller dimension $d_1 \ll qd$, with $d_1=100$. We then use UMAP~\citep{mcinnes2018umap} to further reduce the dimension to 2 for visualization.
\section{Experimental Settings}
\label{sec:experiments}

\begin{figure*}[tbh!]
  \centering
  \includegraphics[width=0.85\linewidth]{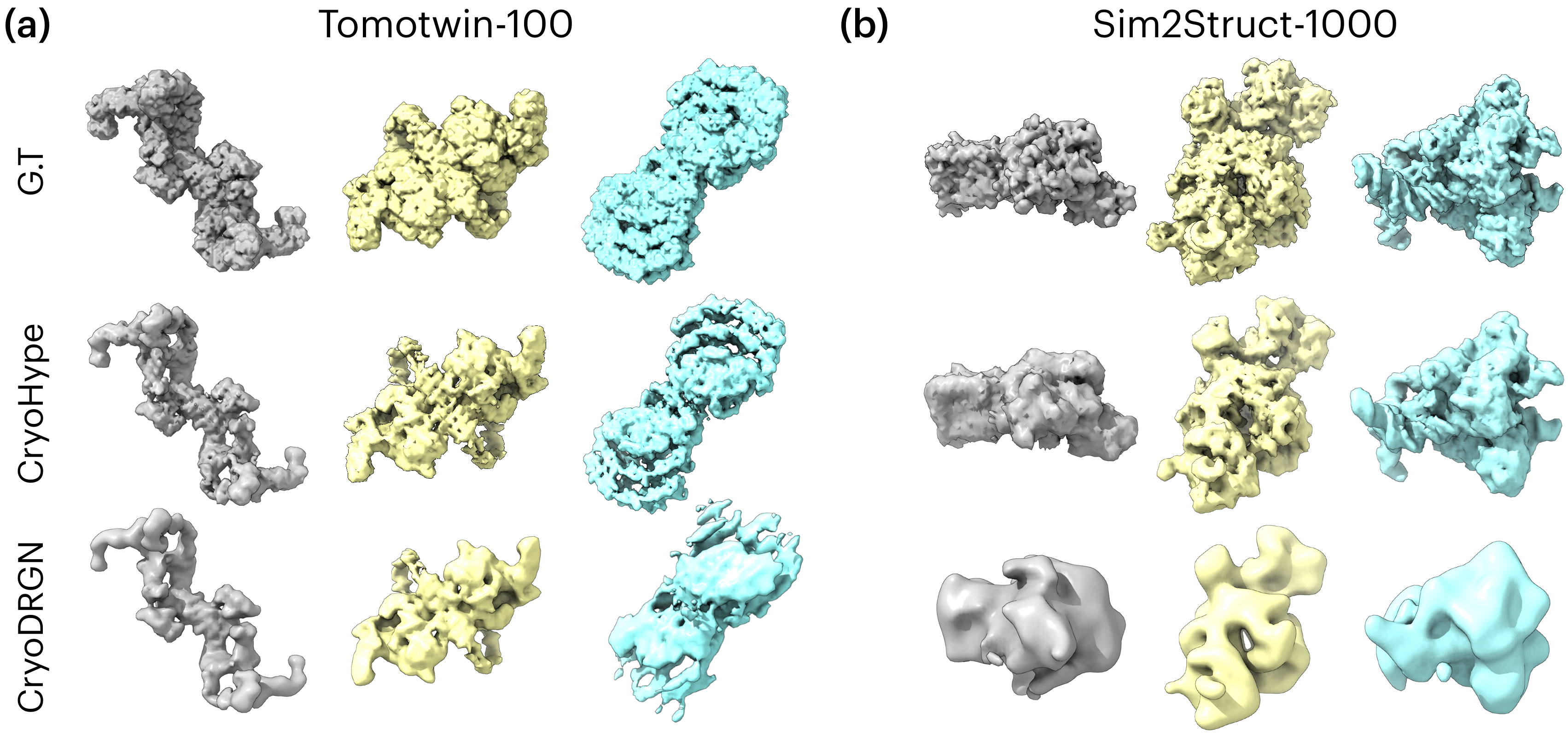}
  \caption{\textbf{Qualitative results for }\texttt{Tomotwin-100} \textbf{and} \texttt{\dataset/}. Representative density volumes and the corresponding ground truth volume. Additional examples are given in Figure~\ref{fig:si_tt100} and Figure~\ref{fig:si_sim2struct1000} in the Appendix.}
  \label{fig:tomo_sim2struct}
  \vspace{-5pt}
\end{figure*}

\subsection{Datasets}
We evaluate our method on two heterogeneous synthetic datasets containing extreme compositional heterogeneity: \texttt{Tomotwin-100}~\citep{jeon2024cryobench} and our new challenging \texttt{\dataset/} dataset. We further demonstrate our method on EMPIAR-10076, an experimental dataset of the assembling ribosome~\citep{davis2016modular}.



\begin{figure*}
  \centering
  \includegraphics[width=0.95\linewidth]{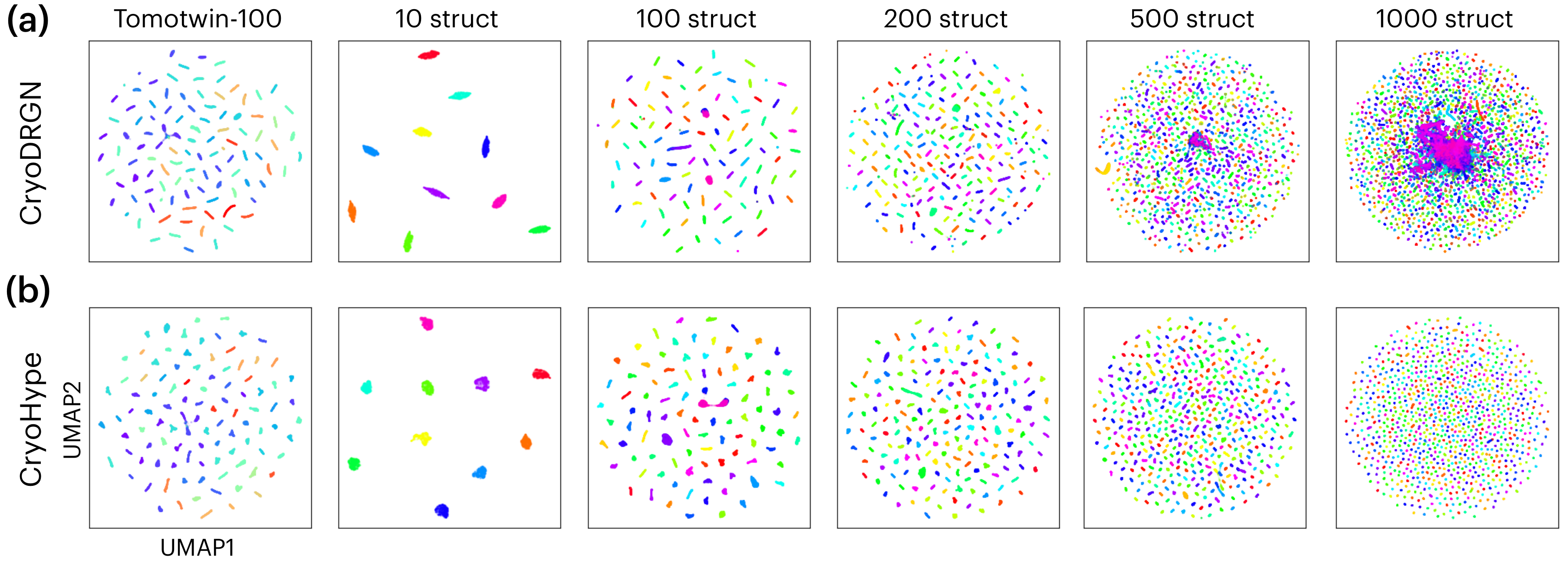}
  \caption{\textbf{Latent visualization for }\texttt{Tomotwin-100} \textbf{and} \texttt{\dataset/}. (a) Latent embeddings from cryoDRGN visualized by UMAP and colored by the 10, 100, 200, 500, and 1000 G.T proteins. (b) Latent embeddings for CryoHype.}
  \label{fig:latent_umap}
  \vspace{-10pt}
\end{figure*}

\begin{table*}
  \centering
  \footnotesize
  \caption{\textbf{Quantitative performance on Tomotwin-100}, measured by $\mathrm{FSC}_\mathrm{AUC}$, CD, and vIoU. $\dagger$ results are from \cite{jeon2024cryobench}. Metrics are also computed on backprojected images for each G.T. structure as an upper bound.}
  \resizebox{0.9\textwidth}{!}{
      \begin{tabular}{c|c c|c c|c c}
        \toprule
        
        \multirow{2}{*}{\textbf{Method}} & \multicolumn{6}{|c}{\textbf{Tomotwin-100}}  \\
        \cmidrule(r){2-7}
        & $\uparrow$ Mean $\mathrm{FSC}_{\mathrm{AUC}}$ (std) & Median & $\downarrow$ Mean CD (std) & Med & $\uparrow$ Mean vIoU (std) & Med \\
        \midrule
        CryoDRGN \citep{zhong2021cryodrgn} & 0.316 (0.046)$^\dagger$ & 0.321$^\dagger$ & \underline{2.26 (1.59)} & \textbf{1.98} & \textbf{0.63 (0.08)} & \textbf{0.65} \\
        DRGN-AI-fixed \citep{levy2025cryodrgn}& 0.202 (0.044)$^\dagger$ & 0.207$^\dagger$ & 32.60 (18.45) & 29.52 & 0.13 (0.09) & 0.12 \\
        Opus-DSD \citep{luo2023opus} & 0.237 (0.049)$^\dagger$ & 0.251$^\dagger$ & 33.48 (0.1378) & 28.92 & 0.14 (0.08) & 0.13 \\
        SFBP \citep{kimanius2022sparse} & 0.036 (0.011) & 0.036 & 18.52 (8.33) & 17.32 & 0.16 (0.06) & 0.16 \\
        3DVA \citep{punjani20213d} & 0.088 (0.040)$^\dagger$ & 0.077$^\dagger$ & 25.52 (17.90) & 21.40 & 0.18 (0.09) & 0.18 \\
        RECOVAR \citep{gilles2025cryo} & 0.258 (0.109)$^\dagger$ & 0.254$^\dagger$ & 27.22 (18.86) & 23.14 & 0.16 (0.08) & 0.15 \\
        3D Class \citep{punjani2017cryosparc} & 0.046 (0.026)$^\dagger$ & 0.037$^\dagger$ & - & - & - & - \\
        \midrule
        \textbf{CryoHype} & \textbf{0.346 (0.033)} & \textbf{0.353} & \textbf{2.18 (0.46)} & \underline{2.11} & \underline{0.61 (0.06)} & \underline{0.62} \\
        \midrule
        Backprojection & 0.364 (0.023) & 0.364 & 1.50 (0.20) & 1.50 & 0.71 (0.03) & 0.71 \\
        \bottomrule
      \end{tabular}
      \vspace{-10pt}
    }
    \label{tab:tab_per-img_snr001}
\end{table*}

\textbf{CryoBench~\citep{jeon2024cryobench}.} 
We evaluate on two synthetic datasets from CryoBench: \texttt{IgG-1D} and \texttt{Tomotwin-100}. The \texttt{IgG-1D} dataset simulates a 1D circular motion of a fragment antibody (Fab) domain of the human immunoglobulin G (IgG) antibody complex. \texttt{IgG-1D} is intended as a simple, diagnostic dataset and is used to validate our approach on conformational heterogeneity (Appendix~\ref{sec:si_confhet}). \texttt{Tomotwin-100} evaluates the capability of cryo-EM reconstruction algorithms to address extreme compositional heterogeneity. This dataset was generated by simulating the cryo-EM image formation process for 100 of the 120 distinct cellular complexes included in the TomoTwin dataset~\citep{rice2023tomotwin}, curated to contain diverse and dissimilar proteins. Notably, \texttt{Tomotwin-100} represents the most challenging dataset in~\citep{jeon2024cryobench}, with most methods failing to achieve successful reconstructions. 

\textbf{\dataset/.}
To evaluate model scalability to datasets containing a large degree of compositional heterogeneity, we introduce \texttt{\dataset/}, a large-scale simulated cryo-EM dataset derived from the Cryo2StructData collection~\citep{giri2024cryo2structdata}. Cryo2StructData comprises experimentally obtained cryo-EM density maps paired with atomic models from the Protein Data Bank (PDB)~\citep{berman2000protein}. Experimental cryo-EM maps from the original collection exhibited inconsistent resolution, noise levels, and grid dimensions due to diverse experimental parameters, potentially introducing confounding downstream biases. To avoid training models that learn these experimental settings, we instead selected a subset of 1000 atomic models filtered by particle size for \texttt{\dataset/}. Each atomic model was converted to a density map and subsequently projected to create 1000 simulated images ($256\times256$, 3.0 Å/pix, downsampled to $128 \times 128$), resulting in a dataset of 1M total particle images (Figure~\ref{fig:ambris-dataset}). \texttt{\dataset/} thus allows evaluation of our method’s robustness under challenging conditions of compositional heterogeneity at scale.
In our experiments, we examine four subsets of this dataset, representing different amounts of compositional heterogeneity, consisting of 10, 100, 200, 500, and all 1000 structures. Each structure has 1000 simulated projection images. The complete details of \dataset/'s construction can be found in Appendix~\ref{si_sim2struct}.

\textbf{EMPIAR-10076.}
We also evaluate our method on an experimental dataset, \texttt{EMPIAR-10076}~\citep{davis2016modular}, which is known to exhibit significant compositional heterogeneity, comprising 14 discrete structures of the assembling 50S ribosome organized into four major assembly states. The data is preprocessed according to~\cite{zhong2021cryodrgn}.

\subsection{Metrics}
\label{sec:metrics}

We measure reconstruction quality with three metrics. The first, Fourier Shell Correlation (FSC), is a standard metric for comparing volumes in cryo-EM, computing correlation between Fourier shells at various thresholds and is a global measure of resolution, but can be misleading in the heterogeneous case \citep{gilles2025cryo}. We follow~\citep{jeon2024cryobench} and evaluate methods using the area under the FSC curve per image ($\mathrm{FSC}_{\mathrm{AUC}}$). 

In addition, we compute two metrics from 3D shape analysis that measure reconstruction quality in real space and thus are more sensitive to local structural heterogeneity. The first is volumetric intersection-over-union (IoU), which measures the volumetric overlap between volumes, and the second is Chamfer distance (CD), which captures pointwise differences between point clouds. We convert the voxel-based data to point clouds by extracting the coordinates of occupied voxels above a specified density threshold and scaling these to world coordinates based on voxel size and grid dimensions. 
Further analyses of our new metrics and details on the density threshold selection can be found in Appendix~\ref{sec:si-new-metrics}.

\subsection{Baselines}

In Table~\ref{tab:tab_per-img_snr001}, we examine a variety of state-of-the-art fixed-pose methods for cryo-EM reconstruction, including VAE-based reconstruction algorithms \citep{zhong2021cryodrgn, kimanius2022sparse, luo2023opus}, autodecoder-based algorithms \citep{levy2025cryodrgn}, and non-deep learning based algorithms \citep{punjani2017cryosparc, gilles2025cryo, punjani20213d}. For other experiments, we only compare against cryoDRGN~\citep{zhongreconstructing, zhong2021cryodrgn}, the only baseline method that demonstrates reasonable performance in the case of extreme compositional heterogeneity (see Table~\ref{tab:tab_per-img_snr001} and~\citep{jeon2024cryobench}). CryoDRGN is a VAE whose decoder is a conditional implicit neural representation, conditioned by concatenation with the representation produced by the encoder.   
\section{Results}
\label{sec:results}

\begin{table*}
  \centering
  \footnotesize
  \caption{\textbf{Quantitative performance on \dataset/.} Metrics are computed with standard deviations per method in parentheses. Chamfer distance is given in angstroms (\r{A}). Isosurface levels are set at 220 for all subsets of \texttt{\dataset/}.}
  \resizebox{0.98\textwidth}{!}{
      \begin{tabular}{c|c|c c|c c|c c}
        \toprule   
        \multirow{2}{*}{\textbf{Method}} & \multirow{2}{*}{\textbf{Structures}} & \multicolumn{6}{|c}{\textbf{\dataset/}}  \\
        \cmidrule(r){3-8}
        & & $\uparrow$ Mean $\mathrm{FSC}_{\mathrm{AUC}}$ (std) & Median & $\downarrow$ Mean CD (std) & Median & $\uparrow$ Mean vIoU (std) & Median \\
        \midrule
        CryoDRGN & \multirow{2}{*}{10} & 0.434 (0.012) & 0.437 & 1.9898 (0.3010) & 2.0468 & 0.4853 (0.0524) & 0.4806 \\
        \method/ & & \textbf{0.464 (0.006)} & \textbf{0.465} & \textbf{1.7781 (0.1702)} & \textbf{1.7890} & \textbf{0.5005 (0.0336)} & \textbf{0.4939} \\
        \midrule
        CryoDRGN & \multirow{2}{*}{100} & 0.361 (0.039) & 0.357 & 2.3389 (0.6433) & 2.2417 & 0.4731 (0.0602) & 0.4664 \\
        \method/ & & \textbf{0.409 (0.024)} & \textbf{0.407} & \textbf{1.9916 (0.4040)} & \textbf{1.9488} & \textbf{0.4897 (0.0516)} & \textbf{0.4849} \\
        \midrule
        CryoDRGN & \multirow{2}{*}{200} & 0.334 (0.047) & 0.334 & 2.4428 (1.0553) & 2.2273 & \textbf{0.4765 (0.0673)} & \textbf{0.4766} \\
        \method/ & & \textbf{0.377 (0.028)}  & \textbf{0.375} & \textbf{2.0748 (0.3363)} & \textbf{2.0489} & 0.4726 (0.0484) & 0.4697 \\
        \midrule
        CryoDRGN & \multirow{2}{*}{500} & 0.216 (0.069) & 0.213 & 4.6358 (4.2948) & 3.1548 & 0.3866 (0.1293) & 0.4101 \\
        \method/ & & \textbf{0.305 (0.065)} & \textbf{0.322} & \textbf{2.4069 (0.7773)} & \textbf{2.2336} & \textbf{0.4529 (0.0773)} & \textbf{0.4565} \\
        \midrule
        CryoDRGN & \multirow{2}{*}{1000} & 0.139 (0.054) & 0.140 & 9.0656 (7.6560) & 5.9439 & 0.2647 (0.1406) & 0.2608 \\
        \method/ & & \textbf{0.232 (0.079)} & \textbf{0.216} & \textbf{3.0179 (1.2470)} & \textbf{2.6512} & \textbf{0.4181 (0.1088)} & \textbf{0.4394} \\
        \bottomrule
      \end{tabular}
    }
    \label{tab:metrics_sim2struct}
    \vspace{-10pt}
\end{table*}

\begin{table}
    \caption{Ablation study on CryoHype examining the four main components of the model, evaluated by $\mathrm{FSC}_{\mathrm{AUC}}$.}
    \centering
    \label{tab:ablation-study}
    \scalebox{0.85}{
    \begin{tabular}{l|c|c}
        \toprule
        \multirow{2}{*}{\textbf{Method}} & \multicolumn{2}{|c}{\textbf{Tomotwin-100}} \\
         & Mean (std) & Med \\
        \midrule
        Concatenation & 0.255 (0.076) & 0.286 \\
        \midrule
        U-Net encoder & 0.208 (0.031) & 0.214 \\
        MLP encoder & 0.234 (0.032) & 0.240 \\
        \midrule
        \textbf{\method/} & \textbf{0.346 (0.033)} & \textbf{0.353} \\
        \bottomrule
    \end{tabular}
    }
    \vspace{-15pt}
\end{table}


\subsection{CryoBench}


On \texttt{Tomotwin-100}, we find that \method/ outperforms all other methods in terms of $\mathrm{FSC}_{\mathrm{AUC}}$ and is comparable to cryoDRGN in 3D shape metrics (Table~\ref{tab:tab_per-img_snr001}). Qualitatively, we find that \method/ captures more high resolution details than cryoDRGN (Figure~\ref{fig:tomo_sim2struct}(a)), which is also reflected in the FSC curves (Figure~\ref{fig:per_img_fsc}). We attribute this to the more expressive conditioning of the \method/'s hypernetwork architecture. Both \method/ and cryoDRGN produce clustered latent spaces (Figure~\ref{fig:latent_umap}). \method/ has much less variability, as indicated by smaller standard deviations for all metrics. We also find that higher FSCs do not necessarily result in higher Chamfer Distance or volumetric IoU, indicating that our new metrics are capturing differences in structure that are not being captured by FSC. Additional results on \texttt{IgG-1D} can be found in Appendix~\ref{sec:si_confhet}.

\subsection{\dataset/}

Qualitatively, we find that \method/ successfully reconstructs 1000 cryo-EM structures to a  without any prior knowledge of the structures, while cryoDRGN fails to reconstruct most details and even the correct shape (Figure~\ref{fig:tomo_sim2struct}(b)). Quantitatively, we find that \method/ significantly outperforms cryoDRGN at all levels of compositional heterogeneity (10, 100, 200, 500, and 1000 structures) in all metrics (Table~\ref{tab:metrics_sim2struct}). As shown by the FSC curves, \method/ is higher resolution and captures all frequencies better than cryoDRGN across all levels of compositional heterogeneity (Figure~\ref{fig:per_img_fsc}). \method/'s performance advantage over cryoDRGN increases as the compositional heterogeneity gets more extreme, showing the effect of INR parameter oversharing as heterogeneity increases. This trend is also reflected in the latent spaces (Figure~\ref{fig:latent_umap}). We find that while the latent spaces for both methods look well-clustered at lower levels of heterogeneity, the latent space of cryoDRGN starts to degrade at high levels of heterogeneity (500 structures and 1000 structure), indicating that cryoDRGN can no longer completely resolve the heterogeneity in the dataset.  In contrast, the latent space of \method/ remains clustered by structure, even at the most extreme amounts of compositional heterogeneity.

\begin{figure*}[tbh]
  \centering
  \includegraphics[width=.9\linewidth]{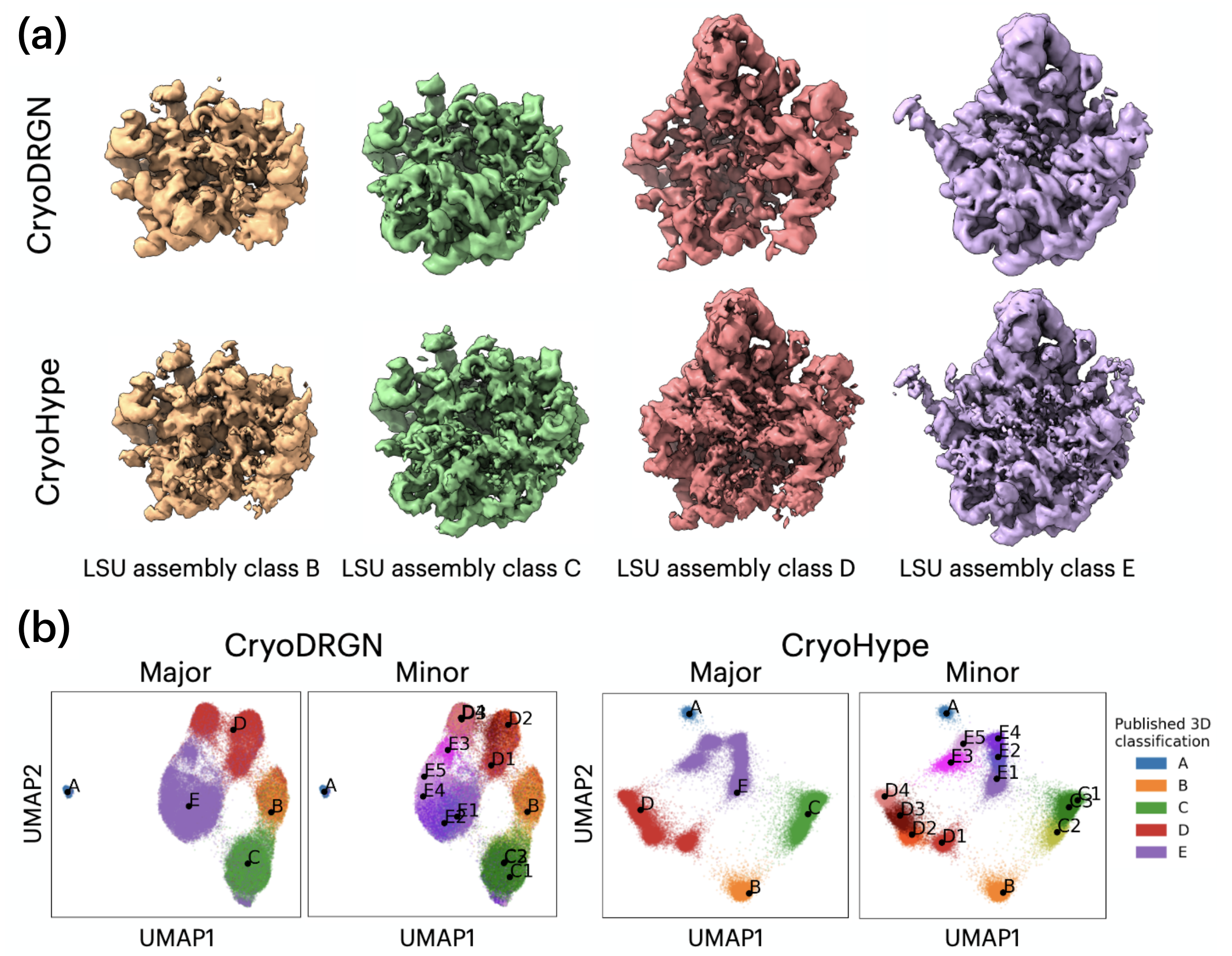}
  \caption{\textbf{Qualitative results on the EMPIAR-10076 dataset.} (a) Density maps of the four major ribosome assembly states from ~\cite{davis2016modular}. 
  (b) Latent space representation, colored by major and minor assembly states assigned from the 3D classification in~\cite{davis2016modular}.} 
  \label{fig:empiar}
  \vspace{-10pt}
\end{figure*}

\subsection{Experimental compositional heterogeneity}

Figure~\ref{fig:empiar}(a) illustrates reconstructed volumes of the four major classes of \texttt{EMPIAR-10076} produced by \method/ and cryoDRGN. Due to the absence of ground truth volumes for this dataset, direct comparison of quality between the methods is challenging, however volumes are qualitatively similar (Fig.~\ref{fig:empiar} (a)). We additionally visualize the latent space colored by major and minor classes identified from the original
publication (Fig.~\ref{fig:empiar} (b)). Both methods successfully separate the major classes, and we note that \method/ produces visually distinct clusters for minor assembly states as well (e.g., D1, D2, D3, and D4).



\subsection{Ablation and analysis}
\label{sec:ablation}

In Table~\ref{tab:ablation-study}, we study the effectiveness of different components of our method by ablating the type of conditioning and the encoder architecture using \texttt{Tomotwin-100}. Hypernetwork conditioning greatly outperforms concatenation, which we attribute to its greater expressivity of (see Section~\ref{sec:methods-motivation}). Changing the encoder from a transformer to a convolutional~\citep{ronneberger2015u} or MLP~\citep{sitzmann2021light} also results in heavily degraded performance despite the convolutional and MLP networks using more parameters, showing the importance of using a ViT encoder in hypernetwork architectures for parameter efficiency and scalability. Additional experiments comparing \method/ against larger cryoDRGN variants and implementation details can be found in Appendix~\ref{sec:si-ablations}. 

\begin{figure}
  \centering
  \includegraphics[width=0.8\linewidth]{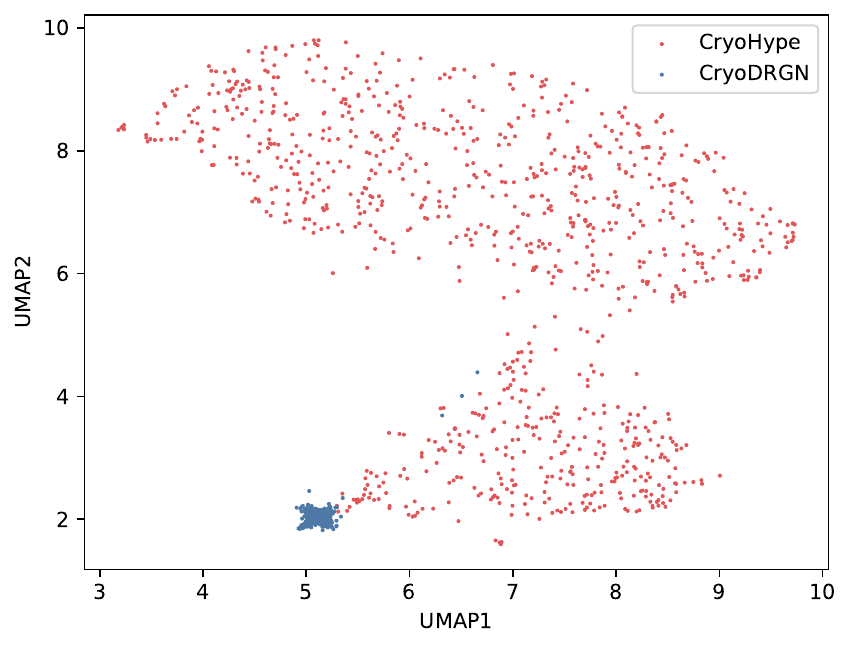}
  \caption{UMAP of activations from the penultimate layer of INRs on \texttt{\dataset/}.}
  \label{fig:activations}
  \vspace{-15pt}
\end{figure}

\begin{figure}
  \centering
  \includegraphics[width=\linewidth]{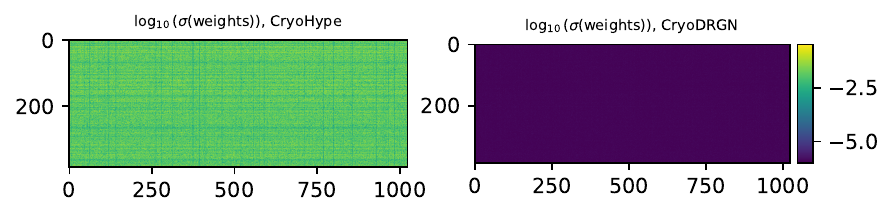}
  \caption{Variation in first-layer INR weights on \texttt{\dataset/}.}
  \label{fig:first-layer-weights}
  \vspace{-18pt}
\end{figure}

We show the reduced parameter-sharing and increased expressivity through two analyses: visualizing the distribution of INR activations (Figure~\ref{fig:activations}) and the distribution of weights in the first INR layer (Figure~\ref{fig:first-layer-weights}). In Figure~\ref{fig:activations}, we plot the activations of INRs generated by \method/ and cryoDRGN on \dataset/ after dimensionality reduction. \method/ generates more diverse activations than cryoDRGN, indicating that its INRs respond differently to the same inputs, a signature of greater expressivity and thus capacity for capturing large-scale compositional heterogeneity. In Figure~\ref{fig:first-layer-weights}, we plot the standard deviations of the weights of the first layer over generated INRs. For cryoDRGN, most weights are effectively constant across INRs, a direct consequence of conditioning by concatenation, whereas almost all weights are modulated for INRs generated by \method/.

\section{Related Work}
\label{sec:related-work}

\textbf{Cryo-EM heterogeneous reconstruction.} 
Current methods for cryo-EM heterogeneous reconstruction can be broadly divided into non-neural and neural network-based approaches
. 3D Classification~\citep{scheres2007disentangling, scheres2012relion, scheres2016processing, punjani2017cryosparc, grant2018cis} uses the Expectation-Maximization algorithm to sort images into a predefined number of discrete classes (typically $< 10$) and is highly sensitive to initialization. Non-neural methods for continuous heterogeneity typically use linear models for heterogeneity~\citep{tagare2015directly, anden2018structural, punjani20213d, gilles2025cryo}. 3DVA~\citep{punjani20213d} and RECOVAR~\citep{gilles2025cryo} are PCA-based methods that use probabilistic PCA and regularized covariance estimation, respectively. These methods learn a linear subspace describing structural heterogeneity, which may be limited in expressivity for modeling compositional heterogeneity arising from diverse mixtures of molecules~\citep{jeon2024cryobench}.

Neural methods typically operate in Fourier space, leveraging the Fourier slice theorem~\citep{bracewell1956strip} to avoid costly numerical integration. CryoDRGN~\citep{zhongreconstructing} and its derivatives~\cite{zhong2021cryodrgn2, levy2022amortized, levy2024mixture} are variational autoencoder~\citep{kingma2013auto} (VAE)-based approaches that use MLP encoders and INR decoders, respectively, while SFBP~\citep{kimanius2022sparse} is a VAE whose decoder is a linear combination of voxel arrays. Opus-DSD~\cite{luo2023opus} is another VAE-based approach that uses a neural volume representation instead of an INR and incorporates a memory bank. Other methods instead represent volumes with a Gaussian Mixture Model~\cite{chen2021deep, schwab2024dynamight}.
3DFlex~\citep{punjani20233dflex}, Hydra~\citep{levy2024mixture}, and cryoDRGN-AI~\citep{levy2025cryodrgn} are encoder-free auto-decoder methods where each structure has a learnable latent code. These existing methods mainly focus on conformational heterogeneity with one or two different species and share almost all of their decoder weights among reconstructed structures.
In contrast, our method aims to model extreme compositional heterogeneity and reduces parameter sharing by dynamically adjusts the weights in every layer of the INR. Our work can also be interpreted as a generalization of discrete mixtures of neural fields~\citep{levy2024mixture} to a continuous mixture of neural fields.

\textbf{Hypernetworks and INRs.} A hypernetwork~\citep{ha2016hypernetworks} is a neural network $g_{\phi}$ that produces or modifies the weights of another neural network $f_\theta$, sometimes called the \textit{primary network} or \textit{hyponetwork}, typically an MLP, with the goal of learning the hypernetwork weights $\phi$. This architecture allows the weights of the primary network to be dynamically adapted to different tasks. 
Most forms of INR conditioning are equivalent to having a hypernetwork producing a subset of its weights~\citep{xie2022neural}. In particular, concatenation, the conditioning approach of most neural cryo-EM reconstruction methods, is equivalent to having an affine hypernetwork network that maps latent codes to the biases of the first layer of the hyponetwork~\citep{sitzmann2020metasdf, dumoulin2018feature, mehta2021modulated}. In between the expressivity of full hypernetworks and concatenation are methods that predict feature-wise transformations~\citep{dumoulin2018feature, chan2021pi, mehta2021modulated}, also called FiLM~\citep{perez2018film} conditioning, which predict a per-layer scale and bias. Our method uses a hypernetwork architecture that predicts the weights (but not the biases) of each layer, making it more expressive than conditioning by concatenation or FiLM. Hypernetworks that produce the weights of the primary network directly are difficult to train~\citep{ortiz2023magnitude}, so often the weights of the primary network are modified using a residual learning approach~\citep{chen2022transformers, ortiz2023magnitude}. Hypernetworks have been widely used to condition INRs~\citep{sitzmann2019scene, sitzmann2020implicit, sitzmann2021light, chen2022transformers, gu2023generalizable, kim2023generalizable, lee2024locality, gu2025foundation}, especially generalizable INRs \citep{chen2022transformers, kim2023generalizable, gu2023generalizable, lee2024locality, gu2025foundation}, where they outperform alternative methods of conditioning INRs such as gradient-based meta-learning \citep{tancik2021learned}. Our insight is that these methods are designed to handle extreme compositional heterogeneity in shapes, with our method  adapting \cite{chen2022transformers} to the task of cryo-EM reconstruction.

\textbf{Heterogeneous benchmarks for cryo-EM.} CryoBench~\citep{jeon2024cryobench} is a benchmark for heterogeneous cryo-EM reconstruction that proposes five datasets with varying types of heterogeneity and degrees of difficulty. Among these, \texttt{Tomotwin-100} tackles large-scale compositional heterogeneity with 100 distinct structures. We extend this further by proposing \texttt{\dataset/}, a dataset containing 1000 structures. Additionally, we note that volume-based metrics for cryo-EM typically rely on Fourier shell correlation (FSC) between volumes, which can be misleading for heterogeneous structures~\citep{gilles2025cryo}. Here, we additionally compute two complementary real-space metrics to provide a more complete evaluation of reconstruction quality (See Section~\ref{sec:metrics}).

\section{Conclusion}
\label{sec:conclusion}

We introduce \method/, a novel transformer hypernetwork approach that can reconstruct datasets with extreme compositional heterogeneity at high resolution. We show that \method/ more accurately recovers compositional heterogeneity from large-scale datasets over previous methods and can produce structured latent spaces by analyzing the parameter space of the INRs. We also introduce \texttt{\dataset/}, a new dataset for compositional heterogeneity, as well as two complementary real-space metrics for evaluating cryo-EM reconstruction quality. 

In this work, we focus on the architectural expressivity of hypernetworks for modeling extreme-scale compositional heterogeneity, and we note that \method/ currently requires known particle poses. While this assumption is unrealistic in real experimental settings, it allows us to isolate and study the benefits of transformer-based hypernetwork conditioning. Extending \method/ to \textit{ab initio} reconstruction with joint pose estimation is an important next step, with natural integration into existing pose-search or amortized frameworks. Beyond pose inference, future work could investigate datasets containing both conformational and compositional heterogeneity, motion recovery within the latent space, and multi-view extensions such as tilt series imaging in cryo-ET~\cite{rangan2024cryodrgn}. Together, these advances suggest that transformer-based hypernetworks, coupled with large-scale heterogeneous datasets, offer a foundation for developing computational methods to enable reconstructing diverse mixtures from cryo-EM at scale.

\section*{Acknowledgments}

The authors would like to thank Anita Rau, Elaine Sui, Jeffrey Heo, Yuhui Zhang, Shiye Su, Orr Zohar, Robert Heeter, Ziyu Xiong, Alkin Kaz, and Yichi Zhang for helpful discussions. The authors acknowledge the use of computing resources at Princeton Research Computing, a
consortium of groups led by the Princeton Institute for Computational Science and Engineering
(PICSciE) and Office of Information Technology's Research Computing. The Zhong lab is grateful
for support from the Princeton Catalysis Initiative, Princeton School of Engineering and Applied
Sciences, the Chan Zuckerberg Initiative, Janssen Pharmaceuticals, Schmidt Sciences, Generate Biomedicines, and NIH grant DP2GM164606.
The funders had no role in study design, data collection and analysis, decision to publish or preparation
of the manuscript.
{
    \small
    \bibliographystyle{ieeenat_fullname}
    \bibliography{main}
}

\clearpage

\appendix
\addcontentsline{toc}{section}{appendices}
\part{Appendix}
\parttoc

\section{Cryo-EM image formation model}
\label{sec:si-image-formation}

\begin{figure*}
  \centering
  \includegraphics[width=0.8\linewidth]{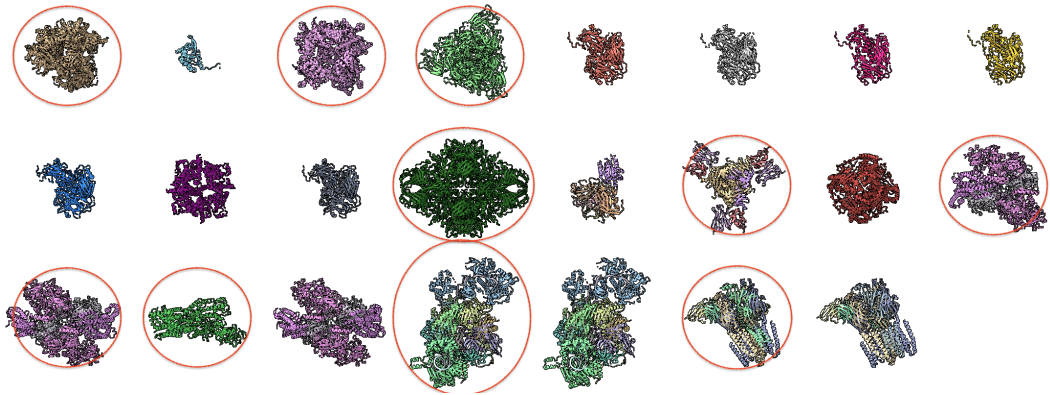}
  \caption{\textbf{Qualitative results of \dataset/ filtering.} Examples of PDB structures from the Cryo2Struct dataset \cite{giri2024cryo2structdata}, highlighting those selected (circled) after filtering based on size and structural distinctiveness.}
  \label{fig:si_filtering}
\end{figure*}

In this section, we expand on Section~\ref{sec:image-formation}. The real space image formation model is not generally used during the training of neural reconstruction methods due to the cost of computing the projection as an integral. Instead, most reconstruction methods exploit the Fourier Slice Theorem (FST)~\cite{bracewell1956strip} to simulate the image formation model efficiently in Fourier space. The FST states that the Fourier transform of a 2D projection of a density volume $V$ is a 2D slice through the origin of $V$ of the 3D Fourier transform of $V$. The FST simplifies the image formation model by circumventing the need to do projection by computing an integral. The Fourier image formation model (Equation~\ref{eqn:real-img-form}) is derived from the real image formation model using the FST.

\section{Architecture and training details}

Here, we provide architectural and training details for our experiments. 

\paragraph{Transformer encoder architecture}
For the \texttt{Tomotwin-100} experiment, our transformer encoder consists of six transformer blocks, with each block consisting of a multi-head self-attention layer followed by an MLP consisting of two linear layers with GeLU activations~\cite{hendrycks2016gaussian}. Each MHSA layer consisted of 12 heads with each head having dimension 64, with the total dimension being 768. The hidden dimension of each MLP is 3072. The Transformer encoder used for the \texttt{\dataset/} experiments were identical except eight blocks were used instead of six.

\paragraph{INR decoder architecture}
For all experiments, \method/ used an INR consisting of 5 linear layers with hidden dimension 1024, random Fourier feature positional encoding, and residual connections between each layer. For fair comparison, these INR architecture hyperparameters were used for both \method/ and all baseline models.

\paragraph{Training hyperparameters}

The \texttt{Tomotwin-100} experiment was carried out with a a batch size of 64, learning rate of 1e-4, a cosine learning rate schedule with a linear warmup of 5 epochs, patch size of 16, and was trained for a total of 50 epochs. For the 10 structure subset of \texttt{\dataset/}, we used a learning rate of 5e-4, while for experiments with more structures, we used a learning rate of 2e-4. All other hyperparameters were the same as the \texttt{Tomotwin-100} experiment. Finally, for the \texttt{EMPIAR-10076}, we used a patch size of 4 with a Gaussian low-pass filter cutoff of 50. A Gaussian low-pass filter is applied to each input token for the Vision Transformer (ViT), with all other hyperparameters being the same as the \texttt{Tomotwin-100} experiment. Hyperparameters were tuned using grid search. 

\paragraph{CryoDRGN}
We trained cryoDRGN using the official PyTorch implementation\footnote{\url{https://github.com/ml-struct-bio/cryodrgn}} (version 3.4.0b). For the synthetic datasets, all results were obtained using the default settings, with the z-dimension set to 8 and the total number of training epochs is 20 as described in~\cite{jeon2024cryobench}. For experiments on \texttt{\dataset/}, we used a batch size of 64 and 50 training epochs. For experiments on \texttt{EMPIAR-10076}, we followed the settings outlined in~\cite{zhong2021cryodrgn}, using 50 training epochs with the latent z-dimension set to 10.

\paragraph{Training splits} As is standard for cryo-EM reconstruction, for all datasets we did not split any of the datasets and used the entire dataset for training. Datasets from CryoBench~\citep{jeon2024cryobench}, including \texttt{Tomotwin-100}, can be downloaded from Zenodo\footnote{\texttt{IgG-1D}: \url{https://zenodo.org/records/12528292}, \texttt{Tomotwin-100}: \url{https://zenodo.org/records/12528292}}.

\paragraph{GPUs, Memory, and Compute Time}

Model were trained on either NVIDIA A100, NVIDIA V100, NVIDIA A6000 GPU, or NVIDIA L40S GPUs. Each model was trained using 2 GPUs using PyTorch Lightning~\cite{PyTorch_Lightning_2024}. Training a \method/ model for 50 epochs and performing inference took approximately 8 hours on 2 A100 GPUs for a dataset with 100K total particles (e.g. \texttt{Tomotwin-100} or the 100 structure subset of \texttt{\dataset/}). \texttt{\dataset/} used 77.7GB CPU memory and uses 24.5GB VRAM for a batch size of 32. For larger datasets, CPU memory consumption can be reduced using lazy dataset loading, at the cost of slower computation. 

\section{Additional details on \dataset/}
\label{si_sim2struct}
We constructed our benchmark, \dataset/, using the Cryo2Struct dataset, initially containing 7,600 cryo-EM density maps paired with corresponding PDB structures \cite{giri2024cryo2structdata}. We specifically chose Cryo2Struct over databases such as AlphaFold DB because each PDB structure is paired with an experimentally determined cryo-EM map, thereby avoiding potential issues associated with synthetic proteins, such as disordered regions complicating downstream structure determination \cite{Abramson2024alphafold3}.

We filtered the initial dataset based on the axis-aligned bounding box dimensions of each PDB structure, retaining only those whose maximum side length fit within the interval $[88, 118)$. This interval ensures each protein comfortably fits within a 256-pixel reconstruction grid, approximately double the protein's maximal length, to accommodate translations within a $\pm20$-pixel range during image projection. This step was critical, as overly small proteins lacked distinctive structural features at our target resolution and too-large proteins would be truncated during image projection. After filtering, we further refined the dataset by eliminating near-duplicate structures, retaining only the first instance of structures sharing identical initial three-character prefixes from their four-character PDB identifiers (e.g. retaining \texttt{6cs3.pdb} out of \texttt{6cs3.pdb}, \texttt{6cs4.pdb}, \texttt{6cs5.pdb}, etc.). Figure~\ref{fig:si_filtering} illustrates the filtering process. From the resulting structures, we selected the first 1000 for further processing, subsequently creating smaller subsets (10, 100, 200, and 500 structures) to evaluate model performance at varying dataset scales. 

Each retained PDB structure was centered by translating the atomic coordinates to place the geometric centroid at the origin. Next, standardized density maps were generated using the \texttt{molmap} command in ChimeraX with parameters set at 3 Å resolution, 1.5 Å grid spacing, and a 256-pixel box size. Although the Cryo2Struct dataset included original EMDB maps, variations in their resolution (ranging 1–4Å) and box dimensions prompted us to generate standardized synthetic volumes to ensure downstream data consistency \cite{giri2024cryo2structdata}.

From these standardized volumes, we simulated cryo-EM images by generating 1000 projections per structure, applying a contrast transfer function (CTF), introducing noise corresponding to an SNR of 0.01, and downsampling images to 128×128 pixels. The final standardized images have dimensions of 128 pixels, 6 Å resolution, and pixel size of 3 Å.

Future work includes evaluating additional protein structure databases to determine their suitability for simulating cryo-EM datasets, particularly to facilitate the development of more generalizable and large-scale cryo-EM reconstruction methods. We plan to release our dataset upon publication on Zenodo\footnote{\url{https://zenodo.org/uploads/17916258}} with the CC-BY license. A backup copy of the dataset will temporarily be available on Dropbox\footnote{\url{https://www.dropbox.com/home/Jeffrey\%20Gu/Sim2Struct}}.

\begin{figure*}
  \centering
  \includegraphics[width=\linewidth]{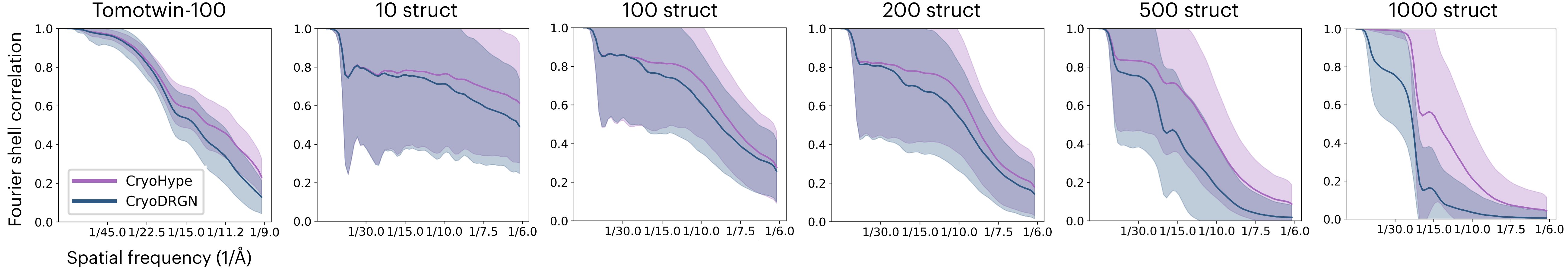}
  \caption{\textbf{Per-Image FSC.} Each curve shows the average FSC curve across all conformations with error bars indicating the standard deviation. The full FSC curves are shown in Appendix.}
  \label{fig:per_img_fsc}
\end{figure*}

\section{Volume metrics for heterogeneous reconstruction}
\label{sec:si-new-metrics}

\paragraph{Per-Image FSC}
We use \textit{per-image FSC} to jointly assess heterogeneity and reconstruction quality following~\cite{jeon2024cryobench}. In cryo-EM, using the Fourier Shell Correlation (FSC) curve is a standard technique for comparing two volumes. The FSC curve calculates the correlation between two volumes (e.g.  a reconstructed volume and a ground truth volume) across spherically averaged radial shells in the Fourier domain. Cryo-EM reconstruction methods often reconstruct a volume from a single input image. \textit{Per-image FSC} evaluates the heterogeneous reconstruction quality of a method by computing the FSC between these per-image reconstructions and ground truth volumes. As in~\cite{jeon2024cryobench}, to compute the per-image FSC we sample one image per conformation to assess the distribution of reconstructions and compute the area under the FSC curve as a summary statistic.

The averaged FSC curves for cryoDRGN and CryoHype on the \texttt{Tomotwin-100} and all subsets of the \texttt{\dataset/} dataset are presented in Figure~\ref{fig:per_img_fsc} and summarized in Table~\ref{tab:tab_per-img_snr001}.

\begin{table*}
    \caption{Ablation study on CryoHype examining the two main components of the model, evaluated by Chamfer distance and volumetric IoU.}
    \centering
    \scalebox{0.95}{
    \begin{tabular}{l|c|c|c|c}
        \toprule
        \multirow{2}{*}{\textbf{Method}} & \multicolumn{4}{|c}{\textbf{Tomotwin-100}} \\
         & $\downarrow$ CD (std) & Med & $\uparrow$ vIoU (std) & Med \\
        \midrule
        Concatenation & 7.157 (11.300) & 2.725 & 0.487 (0.198) & 0.556 \\
        U-Net encoder & 5.154 (2.360) & 4.549 & 0.451 (0.081) & 0.455 \\
        \midrule
        \textbf{\method/} & \textbf{2.185 (0.462)} & \textbf{2.111} & \textbf{0.615 (0.061)} & \textbf{0.621}\\
        \bottomrule
    \end{tabular}
    }
    \label{tab: appendix-ablation-3d-recon}
\end{table*}

\begin{table*}[htb]
    \caption{Comparison of \method/ vs different-sized variants of CryoDRGN~\cite{zhong2021cryodrgn} on the \texttt{Tomotwin-100} dataset, evaluated using $\mathrm{FSC}_{\mathrm{AUC}}$.}
    \centering
    \label{tab:scaling}
    \scalebox{0.85}{
    \begin{tabular}{l|c|c|c}
        \toprule
        \multirow{2}{*}{\textbf{Method}} & \multirow{2}{*}{\textbf{Params}} & \multicolumn{2}{|c}{\textbf{Tomotwin-100}} \\
        & & Mean $\mathrm{FSC}_{\mathrm{AUC}}$ (std) & Med $\mathrm{FSC}_{\mathrm{AUC}}$ \\
        \midrule
        CryoDRGN (base) & 20M & 0.316 (0.046) & 0.321 \\
        CryoDRGN (6 layer encoder) & 21M & 0.316 (0.040) & 0.322 \\
        CryoDRGN (8 layer encoder) & 23M & 0.319 (0.038) & 0.321 \\
        CryoDRGN (6 layer decoder) & 21M & 0.324 (0.040) & 0.328 \\
        CryoDRGN (8 layer decoder) & 23M & 0.324 (0.041) & 0.330 \\
        CryoDRGN (encoder dim 4096, decoder dim 4096) & 155M & 0.338 (0.027) & 0.340 \\
        \midrule
        \textbf{\method/} & \textbf{50M} & \textbf{0.346 (0.033)} & \textbf{0.353} \\
        \bottomrule
    \end{tabular}
    }
\end{table*}

\paragraph{Additional Volume Metrics}
To assess the quality of reconstructed volumes against ground truth volumes, we propose the usage of two new metrics, volumetric IoU and Chamfer distance. \textit{Volumetric IoU} (vIoU) is defined as the intersection of  the ground truth and predicted volume divided by their union. Higher vIoU values indicate better alignment between the predicted and ground truth volumes. We evaluate shape accuracy using \textit{Chamfer Distance} (CD), a metric commonly used for assessing the quality of point cloud reconstructions. CD measures the average bidirectional distance between points in the ground truth and predicted point clouds. Lower CD values indicate more accurate reconstructions. Together, these two metrics provide complementary insights into reconstruction quality by evaluating volumetric overlap and shape precision.

\begin{figure*}
  \centering
  \includegraphics[width=1.0\linewidth]{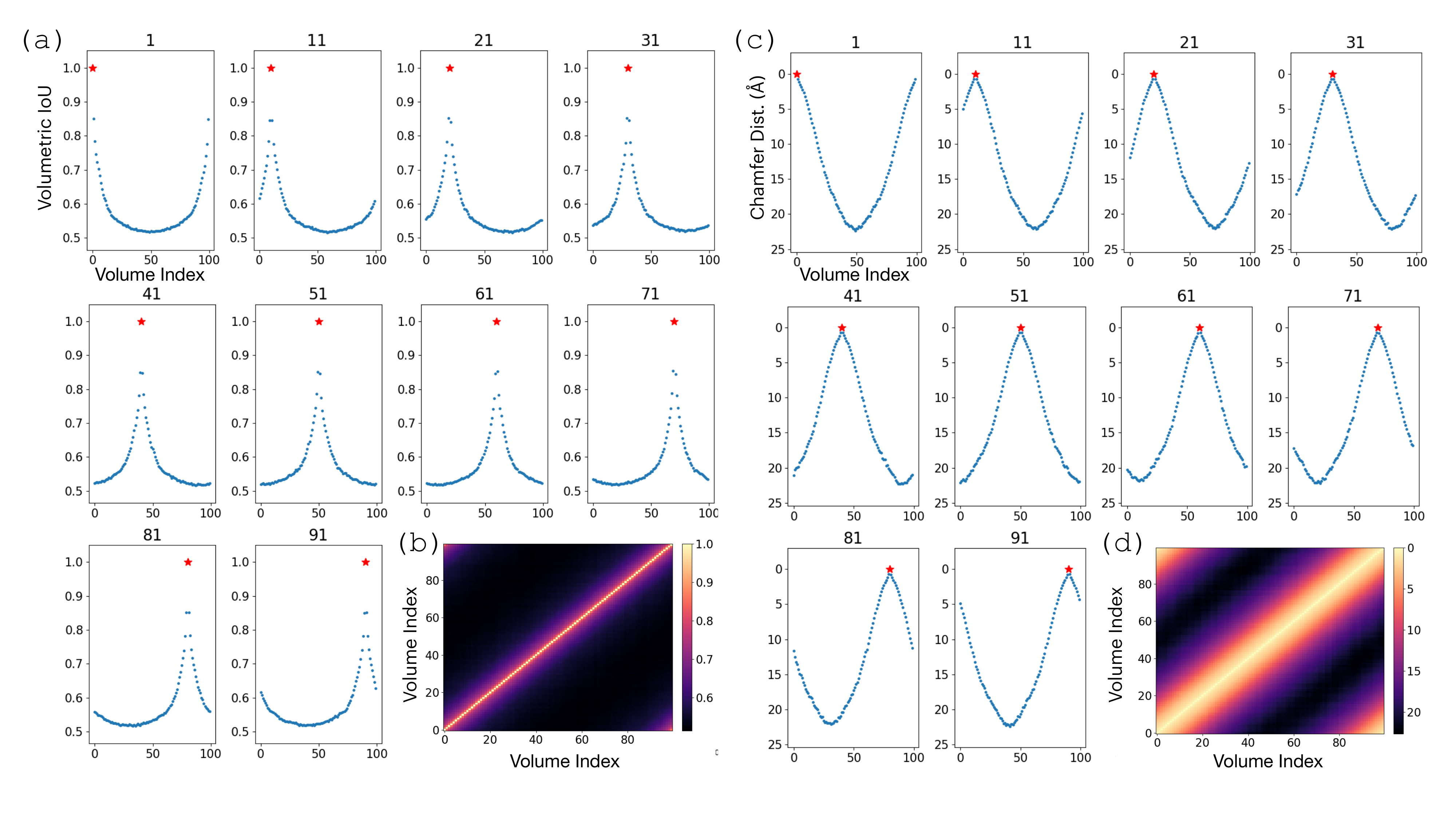}
  \caption{\textbf{Dynamic range of CD and vIoU metrics between ground truth volumes for \texttt{IgG-1D}.} \textbf{(a)} Volumetric IoU computed between one reference G.T. structure and all 100 G.T. structures of the \texttt{IgG-1D} dataset. Each plot corresponds to one reference G.T. volume, indicated by the number above the plot and the red star in each plot. Points higher on the y-axis indicate greater structure similarity. \textbf{(b)} Heatmap of vIoU showing the pairwise volumetric IoU for all pairs of G.T. structures. Lighter colors indicate greater structure similarity. Parts \textbf{(c)} and \textbf{(d)} display the corresponding plots for Chamfer distance.}
  \label{fig:appendix-igg1d}
\end{figure*}

\begin{figure*}
  \centering
  \includegraphics[width=0.5\linewidth]{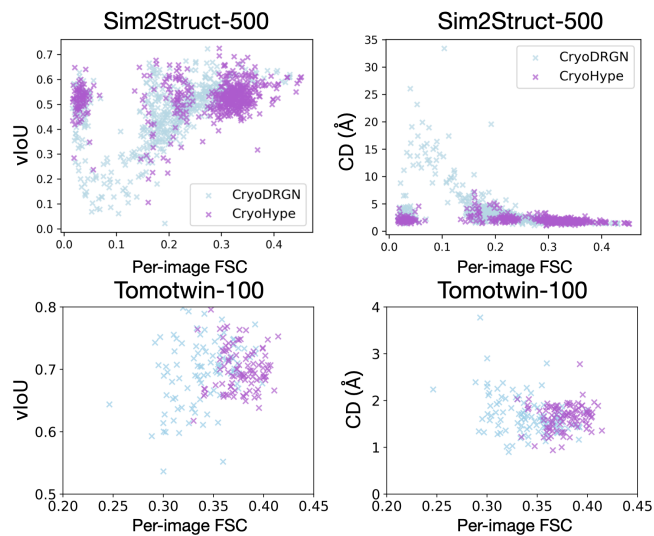}
  \caption{\textbf{Quantitative Metric Comparison for \texttt{Tomotwin-100} and \texttt{\dataset/}.} Higher FSC values positively correlate with vIoU and negatively correlate with CD, reflecting improved reconstruction accuracy. CryoHype (purple) outperforms cryoDRGN (blue) across both datasets.}
  \label{fig:si_scatter}
\end{figure*}

The results in Table~\ref{tab: appendix-ablation-3d-recon} demonstrate the impact of CryoHype’s main components on these metrics, demonstrating their ability to capture insights on performance. Figure~\ref{fig:appendix-igg1d} validates these two metrics on the \texttt{IgG-1D} dataset, a dataset modeling a 1D circular motion of one of the fragment antibody (Fab) domains of the human immunogloblulin G (IgG) protein~\citep{jeon2024cryobench}. Figure~\ref{fig:appendix-violin} highlights metric sensitivity to dataset-specific structural characteristics and the increased difficulty posed by more diverse datasets like \texttt{\dataset/}. 
Figure~\ref{fig:si_scatter} shows the correlation between per-image FSC and our proposed metrics.

\begin{figure*}
  \centering
  \includegraphics[width=0.6\linewidth]{Figures/SI/violin_new.pdf}
  \caption{\textbf{Dynamic range of CD and vIoU metrics between ground truth volumes for each dataset.} Violin plots show the distribution of Chamfer distance (CD) and volumetric IoU (vIoU) metrics on all G.T. structures for four datasets: \texttt{\dataset/}, \texttt{Tomotwin-100}, \texttt{IgG-RL}, and \texttt{IgG-1D}. Lower CD and higher vIoU indicate better shape and volume similarity. \texttt{\dataset/} and \texttt{Tomotwin-100} exhibit wider variation due to greater structural diversity, while \texttt{IgG-1D} shows the most compact distributions, reflecting its homogeneity.}
  \label{fig:appendix-violin}
\end{figure*}

\paragraph{Density Threshold Selection}
To determine which voxels constitute the foreground (object) versus the background in volumetric data, we tuned near-optimal density thresholds to 6.0 for predictions on the \texttt{Tomotwin-100} dataset, 220 for \texttt{\dataset/} predictions, and \(5 \times 10^{-5}\) for all ground truth structures, ensuring consistent voxel segmentation for evaluation. To aid in selecting these thresholds, we opened the figures in ChimeraX~\cite{pettersen2021ucsf} to visually assess and refine the isosurface settings. Final levels were chosen to maximize the completeness of the visualized structures and minimize noisy artifacts. 

\begin{table*}[htb]
    \caption{\method/ and cryoDRGN~\cite{zhong2021cryodrgn} evaluated using traditional FSC metrics on both the \texttt{Tomotwin-100} dataset and all subsets of the \texttt{\dataset/}. Lower is better, with 2 the best possible FSC.}
    \centering
    \label{tab:trad-fsc}
    \scalebox{0.85}{
    \begin{tabular}{c|c|c|c|c|c|c}
        \toprule
        \multirow{2}{*}{\textbf{Method}} & \multirow{2}{*}{\textbf{Dataset}} & \multirow{2}{*}{\textbf{Structures}} & \multicolumn{2}{|c}{\textbf{FSC at 0.143}} & \multicolumn{2}{|c}{\textbf{FSC at 0.5}} \\
        & & & Mean & Median & Mean & Median \\
        \midrule
        CryoDRGN & \multirow{2}{*}
        {Tomotwin-100} & \multirow{2}{*}{100} & 2.17 & 2.06 & 3.14 & 3.08 \\
        \method/ & & & \textbf{2.03} & \textbf{2.00} & \textbf{2.81} & \textbf{2.61} \\
        \midrule
        CryoDRGN & \multirow{2}{*}
        {Sim2Struct-10} & \multirow{2}{*}{10} & \textbf{2.00} & \textbf{2.00} & \textbf{2.00} & \textbf{2.00} \\
        \method/ & & & \textbf{2.00} & \textbf{2.00} & \textbf{2.00} & \textbf{2.00} \\
        \midrule
        CryoDRGN & \multirow{2}{*}
        {Sim2Struct-100} & \multirow{2}{*}{100} & 2.01 & \textbf{2.00} & 2.73 & 2.75 \\
        \method/ & & & \textbf{2.00} & \textbf{2.00} & \textbf{2.30} & \textbf{2.29} \\
        \midrule
        CryoDRGN & \multirow{2}{*}
        {Sim2Struct-200} & \multirow{2}{*}{200} & 2.18 & 2.01 & 3.09 & 2.91 \\
        \method/ & & & \textbf{2.01} & \textbf{2.00} & \textbf{2.63} & \textbf{2.72} \\
        \midrule
        CryoDRGN & \multirow{2}{*}
        {Sim2Struct-500} & \multirow{2}{*}{500} & 3.96 & 3.37 & 5.93 & 4.74 \\
        \method/ & & & \textbf{2.68} & \textbf{2.29} & \textbf{3.49} & \textbf{3.20} \\
        \midrule
        CryoDRGN & \multirow{2}{*}
        {Sim2Struct-1000} & \multirow{2}{*}{1000} & 6.45 & 6.10 & 9.31 & 6.74 \\
        \method/ & & & \textbf{3.87} & \textbf{3.76} & \textbf{4.53} & \textbf{4.57} \\
        \bottomrule
    \end{tabular}
    }
\end{table*}

\paragraph{Thresholding Limitations and Implications}
 A current limitation of our new metrics is that each predicted structure may require individual density level tuning, which limits the ability to draw dataset-wide performance conclusions using these density-dependent metrics. A potential solution to this is to optimize the density level for each predicted structure on a given metric with a greedy search algorithm. Moreover, Chamfer distance is highly sensitive to outliers, resulting in occasional extreme values, particularly at lower density thresholds where noisy or disconnected points are included. Higher density thresholds reduce the number of high-CD outliers but must be carefully tuned to avoid excessive loss of volume and structural details. This sensitivity highlights the importance of selecting density thresholds that balance object completeness with noise reduction.

\section{Ablations}
\label{sec:si-ablations}

\begin{table*}[htb]
    \caption{\method/ and cryoDRGN~\cite{zhong2021cryodrgn} evaluated using supervised classification metrics on all subsets of the Sim2Struct dataset.}
    \centering
    \label{tab:classification}
    \scalebox{0.85}{
    \begin{tabular}{c|c|c|c|c|c|c}
        \toprule
        \textbf{Method} & \textbf{Dataset} & \textbf{Structures} & \textbf{Accuracy} & \textbf{Precision} & \textbf{Recall} & \textbf{F1} \\
        \midrule
        CryoDRGN & \multirow{2}{*}
        {Sim2Struct-10} & \multirow{2}{*}{10} & 0.9998 & 0.9998 & 0.9998 & 0.9998 \\
        \method/ & & & \textbf{0.9999} & \textbf{0.9999} & \textbf{0.9999} & \textbf{0.9999} \\
        \midrule
        CryoDRGN & \multirow{2}{*}
        {Sim2Struct-100} & \multirow{2}{*}{100} & \textbf{0.9621} & \textbf{0.9523} & \textbf{0.9621} & \textbf{0.9555} \\
        \method/ & & & 0.9506 & 0.9421 & 0.9506 & 0.9451 \\
        \midrule
        CryoDRGN & \multirow{2}{*}
        {Sim2Struct-200} & \multirow{2}{*}{200} & 0.9397 & 0.9298 & 0.9397 & 0.9311 \\
        \method/ & & & \textbf{0.9632} & \textbf{0.9557} & \textbf{0.9632} & \textbf{0.9581} \\
        \midrule
        CryoDRGN & \multirow{2}{*}
        {Sim2Struct-500} & \multirow{2}{*}{500} & 0.8375 & 0.8202 & 0.8375 & 0.8166 \\
        \method/ & & & \textbf{0.9719} & \textbf{0.9635} & \textbf{0.9719} & \textbf{0.9663} \\
        \midrule
        CryoDRGN & \multirow{2}{*}
        {Sim2Struct-1000} & \multirow{2}{*}{1000} & 0.6792 & 0.6508 & 0.6792 & 0.6468 \\
        \method/ & & & \textbf{0.9753} & \textbf{0.9713} & \textbf{0.9753} & \textbf{0.9724} \\
        \bottomrule
    \end{tabular}
    }
\end{table*}

\begin{table*}[htb]
    \caption{Performance of cryoSPARC's fixed pose 3D classification method on the Sim2Struct dataset, evaluated using $\mathrm{FSC}_{\mathrm{AUC}}$.}
    \centering
    \label{tab:3dclass}
    \scalebox{0.85}{
    \begin{tabular}{l|l|c|c}
        \toprule
        \multirow{2}{*}{\textbf{Method}} & \multirow{2}{*}{\textbf{Dataset}} & \multicolumn{2}{|c}{$\mathrm{FSC}_{\mathrm{AUC}}$} \\
        & & Mean (std) & Med \\
        \midrule
        cryoSPARC 3D class (fixed pose) & Sim2Struct-10 & 0.204 (0.036) & 0.224 \\
        cryoSPARC 3D class (fixed pose) & Sim2Struct-100 & 0.071 (0.055) & 0.037 \\
        \bottomrule
    \end{tabular}
    }
\end{table*}

In this section, we provide further details on the ablation experiments in Section~\ref{sec:ablation}. Additional quantitative results using our proposed Chamfer distance and volumetric IoU metrics can be found in Table~\ref{tab: appendix-ablation-3d-recon}. We also provide ablations of different sized cryoDRGN~\cite{zhong2021cryodrgn} variants vs CryoHype.

\paragraph{Concatenation ablation}
We modify our architecture to only condition the INR decoder by concatenation, turning our network into an autodecoder. We do this by removing all linear heads $\texttt{Head}_i$ except one (see Sec~\ref{sec:vit-hypernet}) and modifying the forward pass of the network to apply the single linear head to just the last output token to produce a latent vector $\mathbf{z}$, which is used to condition the decoder by concatenation.

\paragraph{U-Net hypernetwork}
The U-Net~\cite{ronneberger2015u, buda2019association} encoder for the U-Net ablation takes as input images of shape $[1, D, D]$ and has output shape $[L, D, D]$, where $L$ is the number of layers of the INR and $D=256$ is a multiple of one of the dimensions of the INR's weight matrices and initial features size 144. Since our input originally has shape $[1, 129, 129]$, we reshape the input to size $[1, D, D]$ using interpolation with nearest upsampling. We then apply each linear head $\texttt{Head}_i, 1 \le i \le L$ to a channel of the output, transforming $D$ to the correct size for the given layer. We then repeat the other dimension as necessary to get the full correct shape. Training is unchanged.

\paragraph{MLP hypernetwork}
We base our MLP hypernetwork architecture on that of Light Field Networks~\cite{sitzmann2021light}, which uses a separate MLP encoder to the generate the weights of each layer. We adapt the architecture of \cite{sitzmann2021light} to produce weight tokens instead of directly producing the weights, leaving the rest of the architecture unchanged from the ViT hypernetwork (see Section~\ref{sec:methods}). 

\paragraph{Larger cryoDRGN variants} The base \method/ model requires more parameters than the base cryoDRGN model (50M vs 20M). In this ablation study, we show that scaling cryoDRGN does not allow it to match the performance of \method/ (Table~\ref{tab:scaling}). While the performance of cryoDRGN increases with increasing encoder and decoder capacity, the performance of cryoDRGN increases only very gradually, showing that \method/ scales much better than cryoDRGN. The largest cryoDRGN variant still shows lower performance than the base \method/ model on Tomotwin-100 (c.f. Table~\ref{tab:tab_per-img_snr001}), despite having 3 times the number of parameters (155M to 50M). 

\section{Additional Quantitative Results}

In this section, we provide additional quantitative evaluations of our method.




\subsection{Traditional FSC}

In Table~\ref{tab:trad-fsc}, we evaluate \method/ vs cryoDRGN~\cite{zhong2021cryodrgn} using the traditional FSC at 0.143 and FSC at 0.5 metrics on both Tomotwin-100 and all \texttt{\dataset/} subsets. We find that, in line with our other quantitative metrics, the performance of \method/ matches or exceeds the performance of cryoDRGN on all datasets, with \method/ performing better as the degree of compositional heterogeneity, as measured by the number of distinct structures, gets more and more extreme. 

\subsection{Supervised Classification Metrics}

Since \method/ was evaluated quantiatively on synthetic datasets in the fixed-pose setting where the particle poses are known, it also makes sense to evaluate the performance of \method/ using supervised classification evaluation metrics. We find that overall, CryoHype outperforms cryoDRGN in supervised classification metrics (Table~\ref{tab:classification}). CryoHype maintains close to perfect classification metrics regardless of the number of structures, whereas cryoDRGN's performance is close to perfect at 10 and 100 structures and begins to drop at 200 structures, degrading rapidly as the number of structures increases. This mirrors the trends observed in the other metrics ($\mathrm{FSC}_{\mathrm{AUC}}$, CD, vIoU). The large drops at 500 and 1000 structures can be seen visually in latent space, where the poor quantitative classification metrics for cryoDRGN are reflected as increasingly poor organization at the center of latent space (see Figure~\ref{fig:latent_umap}).

\subsection{Additional baselines}

In this section, we report the results of traditional maximum likelihood classification methods with fixed pose (e.g. cryoSPARC 3D classification~\cite{punjani2017cryosparc}) on our new \texttt{\dataset/} dataset. As with the cryoSPARC results reported in our paper (originally from CryoBench~\cite{jeon2024cryobench}, see Table~\ref{tab:tab_per-img_snr001}), we confirm that cryoSPARC’s 3D classification method struggles with compositional heterogeneity and especially the extreme compositional heterogeneity considered in our paper (Table~\ref{tab:3dclass}). In particular, the results on the 100 structure subset of \texttt{\dataset/} mirror that of \texttt{Tomotwin-100}. At even more extreme levels of heterogeneity ($>$100 classes), we find that cryoSPARC 3D classification throws an error.

\section{Conformational Heterogeneity}
\label{sec:si_confhet}

\begin{figure}
  \centering
  \includegraphics[width=0.8\linewidth]{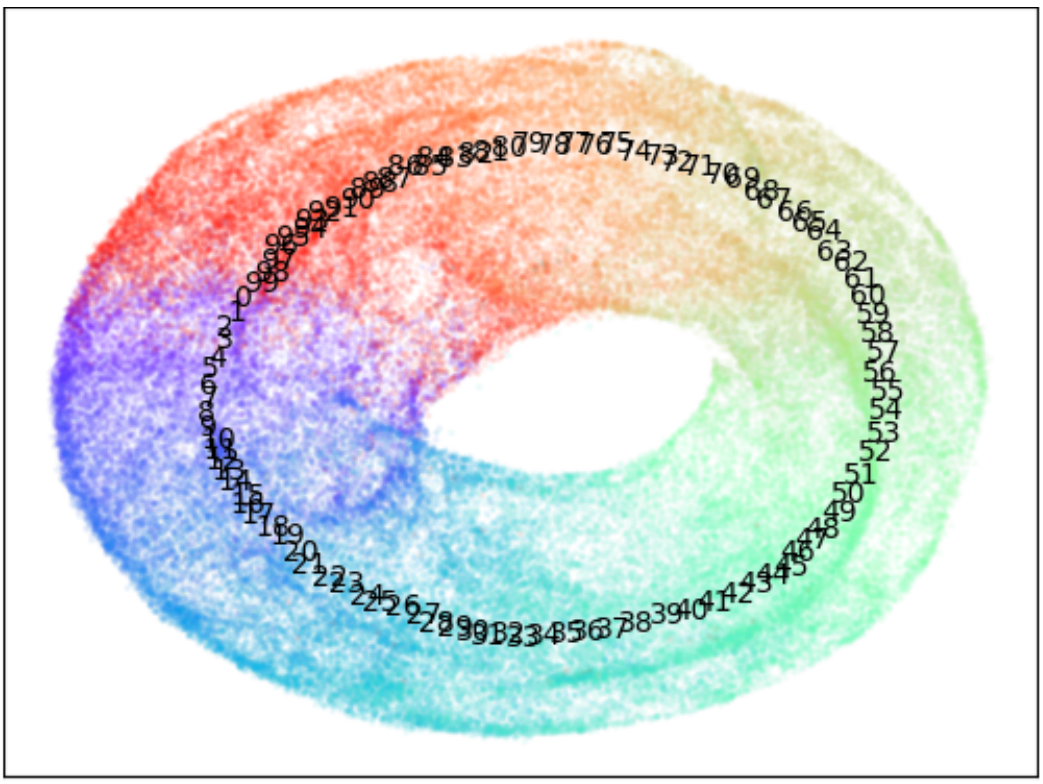}
  \caption{\textbf{IgG-1D latent space} We find that \method/ successfully captures the 1D circular motion of the \texttt{IgG-1D} dataset. The numbers $0 \le i \le 99$ label the mean latent for each of the 100 conformations of \texttt{IgG-1D}, where $(3.6 i)^{\circ}$ is the angle of the Fab domain relative to the fixed domains.}
  \label{fig:si_igg1d_latent}
  \vspace{-18pt}
\end{figure}

Although \method/ was not designed to target conformational heterogeneity, it is still able to recover conformational heterogeneity. In Figure~\ref{fig:si_igg1d_latent}, \method/ is able to successfully recover the circular motion of \texttt{IgG-1D}~\cite{jeon2024cryobench} when augmented with an additional smoothness loss.

\section{Additional Qualitative Results}
In this section, we provide additional examples of reconstructed volumes for both \method/ and cryoDRGN on each dataset. 

\begin{figure*}
  \centering
  \includegraphics[width=0.8\linewidth]{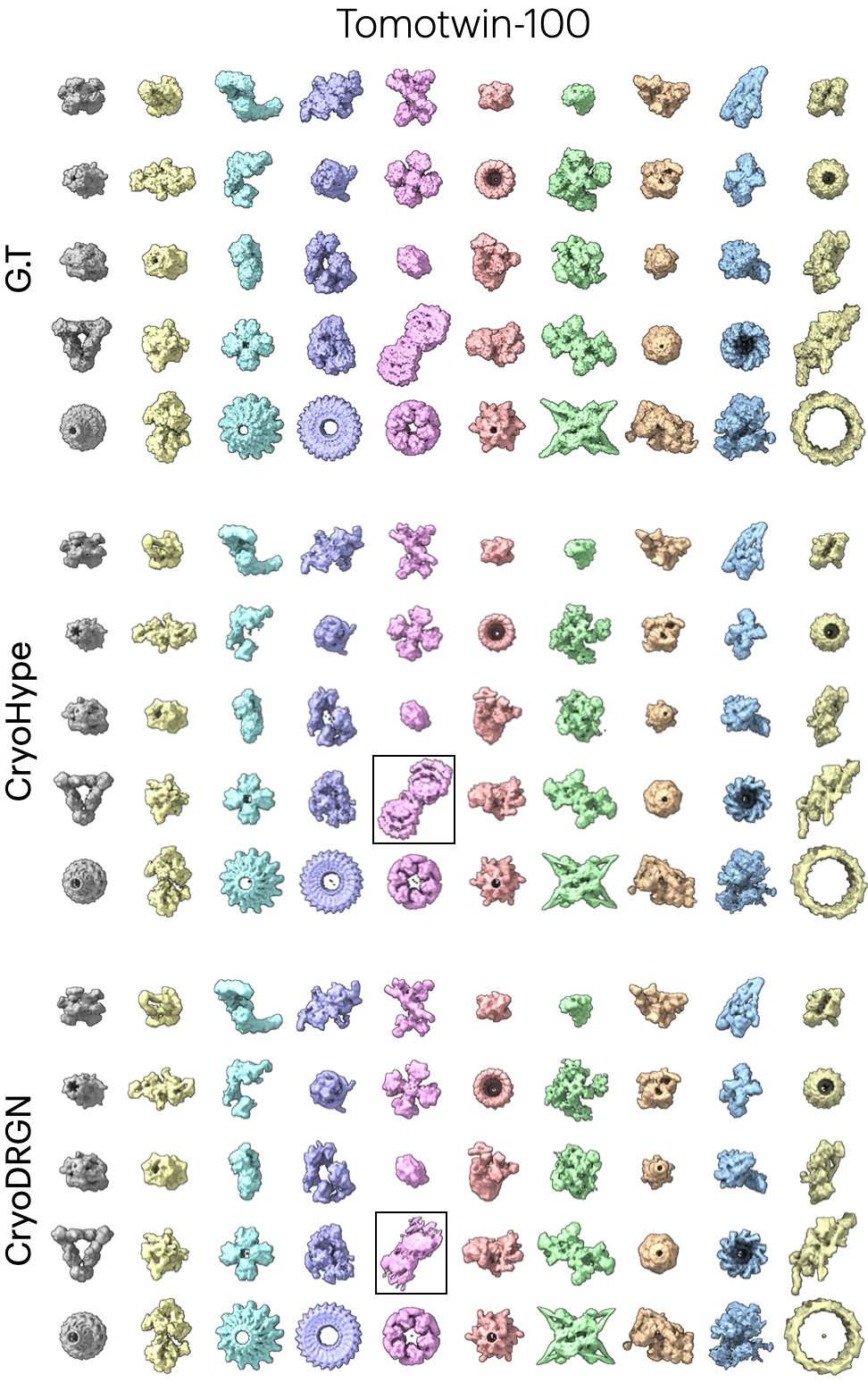}
  \caption{\textbf{Tomotwin-100 (SNR 0.01) qualitative results} Last 50 volumes reconstructed by CryoHype and CryoDRGN for the \texttt{Tomotwin-100} dataset.}
  \label{fig:si_tt100}
\end{figure*}

\begin{figure*}
  \centering
  \includegraphics[width=0.8\linewidth]{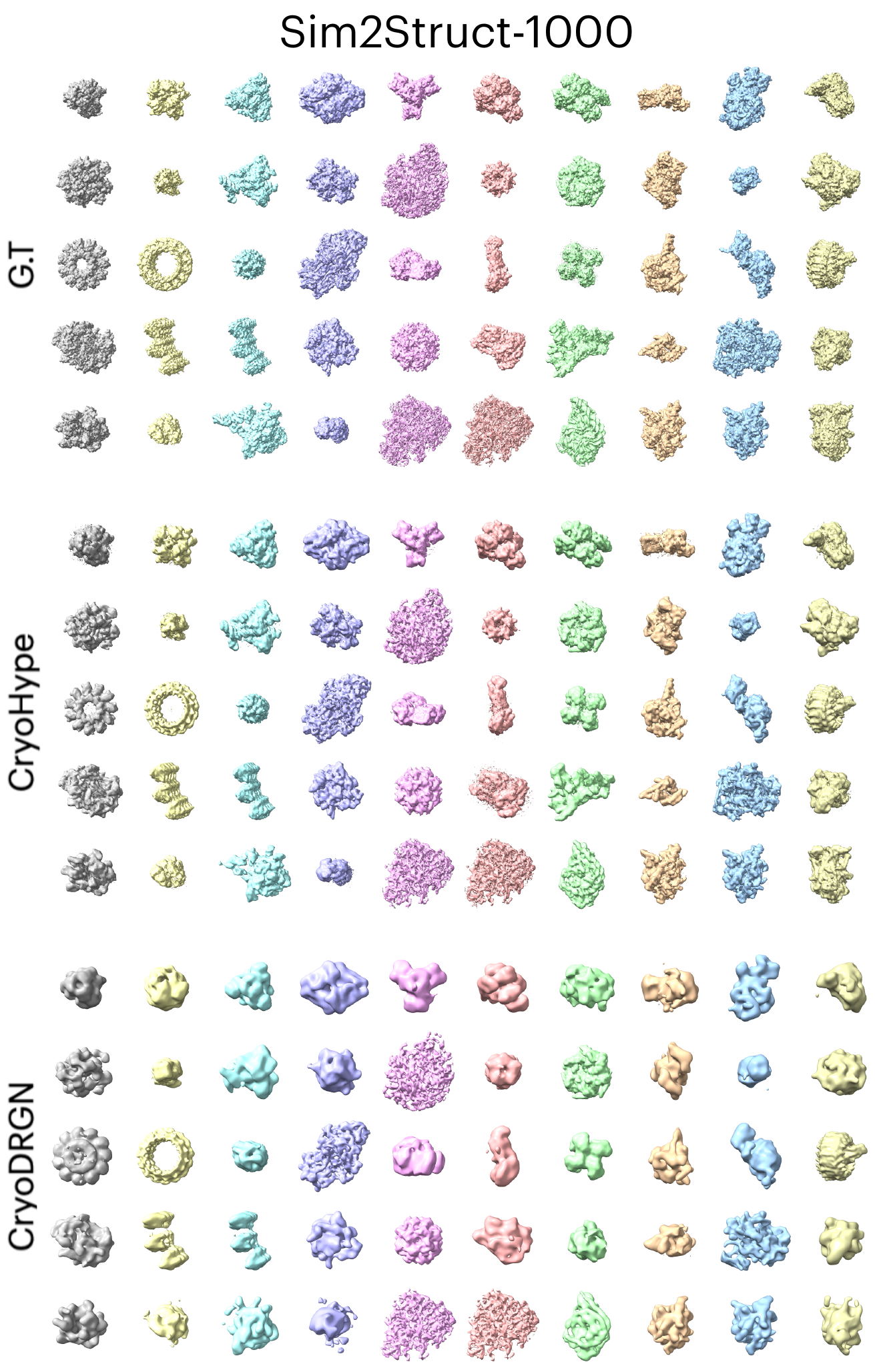}
  \caption{\textbf{Sim2Struct-1000 qualitative results with G.T} Last 50 volumes reconstructed by CryoHype and cryoDRGN for the \texttt{Sim2Struct-1000} dataset.}
  \label{fig:si_sim2struct1000}
\end{figure*}

Figures~\ref{fig:si_tt100} and~\ref{fig:si_sim2struct1000} show the groundtruth, \method/, and cryoDRGN reconstructions for \texttt{Tomotwin-100} and \texttt{\dataset/}, respectively.
We see that overall, \method/ is both able to reconstruct some shapes that cryoDRGN cannot. In particular, cryoDRGN struggles with proteins with loops, while \method/ does a better job of reconstructing these proteins. Additionally, \method/ generally has higher resolution and preserves fine details better than cryoDRGN.  

\begin{figure*}
  \centering
  \includegraphics[width=1.0\linewidth]{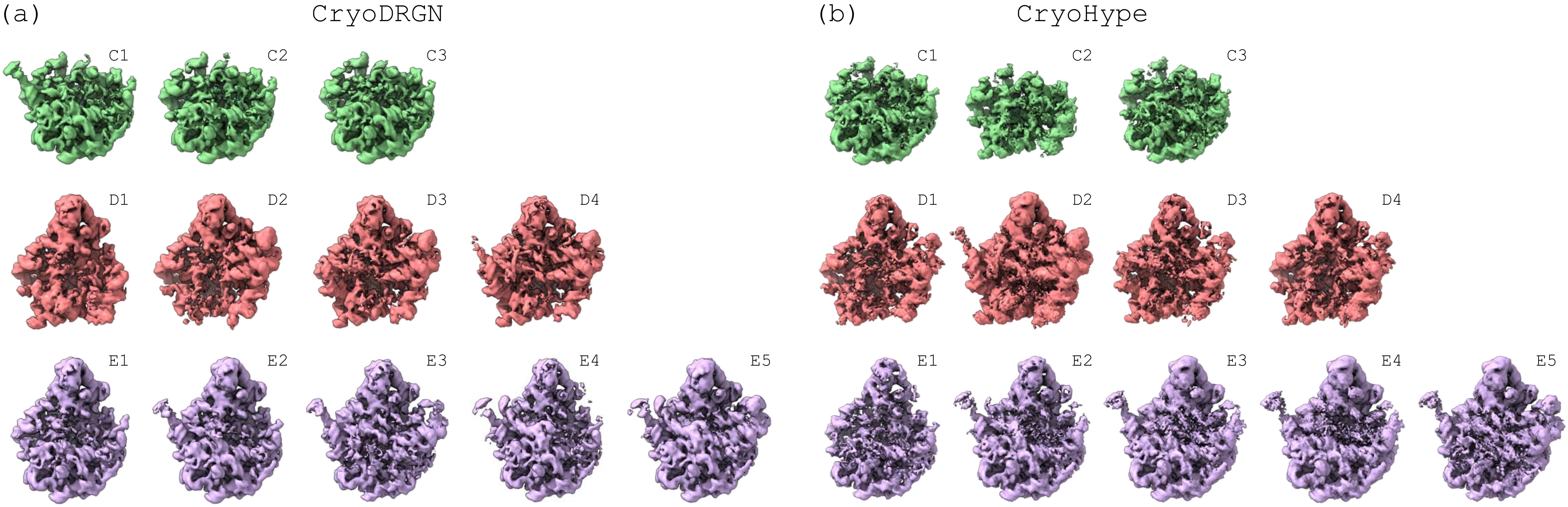}
  \caption{\textbf{EMPIAR-10076} Reconstructed volumes of ribosome assembly minor classes for (a) CryoDRGN and (b) CryoHype.}
  \label{fig:si_empiar_total}
\end{figure*}

Figure~\ref{fig:si_empiar_total} shows reconstructed volumes for each class of \texttt{EMPIAR-10076}. Each CryoHype structure is generated by randomly sampling from latent encoding of particles with the corresponding class assignments from~\cite{davis2016modular}.

\end{document}